\pdfoutput=1

\documentclass[11pt]{article}
\usepackage[table]{xcolor} 
\usepackage[]{acl}

\usepackage{times}
\usepackage{latexsym}
\usepackage{graphicx}
\usepackage{arydshln}
\usepackage{multirow} 
\usepackage{colortbl} 

\usepackage[T1]{fontenc}

\usepackage[utf8]{inputenc}

\usepackage{microtype}

\usepackage{inconsolata}

\usepackage{subcaption}
\title{Do Prompt Positions Really Matter?}

\author{Junyu Mao\thanks{Corresponding author} \and
  Stuart E. Middleton \and
    Mahesan Niranjan \\
  School of Electronics and Computer Science\\
  University of Southampton \\
  \texttt{\{j.mao, sem03\}@soton.ac.uk, \{mn\}@ecs.soton.ac.uk}}

\begin{document}
\maketitle
\begin{abstract}
Prompt-based models have gathered a lot of attention from researchers due to their remarkable advancements in the fields of zero-shot and few-shot learning. Developing an effective prompt template plays a critical role. However, prior studies have mainly focused on prompt vocabulary searching or embedding initialization within a predefined template with the prompt position fixed. In this empirical study, we conduct the most comprehensive analysis to date of prompt position for diverse Natural Language Processing (NLP) tasks. Our findings quantify the substantial impact prompt position has on model performance. We observe that the prompt positions used in prior studies are often sub-optimal, and this observation is consistent even in widely used instruction-tuned models. These findings suggest prompt position optimisation as a valuable research direction to augment prompt engineering methodologies and prompt position-aware instruction tuning as a potential way to build more robust models in the future.


\end{abstract}

\section{Introduction}
 
Recently, \citet{GPT3} have shown the impressive performance of using handcrafted prompts with a frozen language model in zero-shot and few-shot learning, leading to increased interest and activity in prompt engineering within the NLP community \cite{schick2020IPET,gao2020LMBFF,li-liang-2021-prefix}. Prompting (a.k.a prompt-based learning \cite{promptsurveyLiu}) aims to reformat an NLP problem so that it closely matches the format used in the pre-training tasks. To apply prompt-based learning methods effectively, a critical step involves the creation of a prompt template that maximizes performance on the downstream task.

In many previous works, it is common to manually pre-define a template while keeping the prompt position fixed (e.g. prepend the prompt to the input \cite{lester2021power}). These studies often concentrate more on either prompt vocabulary searching \cite{gao2020LMBFF,shin2020autoprompt,PADA} or prompt embedding initialization \cite{P-tuning,gu2022ppt}. However, there has been limited research exploring how different approaches to positioning the prompt sequences can affect the models’ behaviour, despite indications that varying prompt positions may lead to performance difference \cite{Maoposition, wu2022idpg}.

Hence, in this paper, we quantify how much prompt positions matter by evaluating various accessible models on different NLP tasks under few-shot and zero-shot settings.  We comprehensively test a range of prompt position options with many widely used prompt styles (e.g. cloze and prefix) and methods (e.g. gradient-based and gradient-free). Our findings reveal unexpected performance variations among different prompt positions in both zero-shot and few-shot settings. We also discover that instruction-tuned models do not always reduce performance disparities, even though they typically include vocabulary and positional variations in their training templates. Interestingly, we observe that in many cases, the prompt positions used in previously published works show a sub-optimal performance compared to other prompt position choices.  Our choice of zero and few-shot tasks is driven by the observation that prompting methods are particularly useful when training data is limited \cite{promptsurveyLiu}, and this hypothesis is born out by our results which show prompt positions matter more when the available labelled data is limited.

The key contributions of this paper are\footnote{Code available at \url{https://github.com/milliemaoo/prompt-position}.}:
\begin{itemize}
    \item To the best of our knowledge, we are the first comprehensive analysis looking at the impact of prompt position across different methods and prompt styles in both few-shot and zero-shot settings for a variety of NLP tasks.
    \item Empirical results show that prompt positions matter. The positions used in many published works are often sub-optimal choices, with no universally superior prompt position across all tasks. These results suggest prompt position optimisation might be a useful addition to the existing field of prompt engineering and prompt position-aware instruction tuning could be explored to build more robust language models in the future.
\end{itemize}


\section{Related Work}\label{sec: related_work}
\paragraph{Prompt-based learning.}  Many prior works have concentrated on \textbf{Gradient-based} methods within discrete spaces \cite{schick2020IPET,IPET2,gao2020LMBFF} as well as prompting directly in the embedding space. This latter approach uses tunable prompt tokens that are not limited to natural language, which can be either prepended to the input \cite{lester2021power,ptuning2,gu2022ppt} or be inserted in a hybrid template \cite{P-tuning}.  \citet{sun2022blackbox} optimize continuous tokens without using gradients, although this approach is not suitable for APIs like GPT-3 which only allow for text modifications rather than token embeddings. There are also \textbf{Gradient-free} works focusing on in-context learning \cite{GPT3,lu2021fantastically}, chain-of-thought \cite{wei2022chain,zhang2022autocot,yao2023tree}, and instruction generation \cite{prasad2022grips,zhou2022largemodelsarehuman}, especially when instruction tuning \cite{sanh2022multitask,wei2022instruction,chung2022scaling} plays a key role in the steering process of Large Pre-trained LMs. Our paper includes experiments examining the impact of prompt position from both gradient-based and gradient-free perspectives.

\paragraph{Prompt position.} There is limited work that involves prompt position.  \citet{Maoposition} find that the position of a handcrafted prompt (before or after the input) affects model performance, but there's no consensus on which position is best. \citet{wu2022idpg} propose an instance-dependent prompt generation method; meanwhile, they study the effect of inserting a sequence of prompt tokens in different positions based on their proposed method and prompt tuning \cite{lester2021power}. Recently, \citet{dynamicposition} have proposed a dynamic position method that can significantly improve the performance of prompt tuning. They both point out that different positions of prompts will deliver different results with the consideration of only one specific approach to creating the prompt. In this paper, we present the most comprehensive analysis of prompt positions to date and take into account various types of prompts under both zero-shot and few-shot settings.

\section{Method}\label{sec: method}
\subsection{Prompt Style}
Two common styles of prompts are explored in our experiments: \textbf{Cloze style} aims to let LMs fill in the blanks. For example, the input of sentiment classification \textit{"I love this movie"} can be formulated as \textit{"I love this movie. Overall, it was a [Mask] movie."}, and the model will be asked to predict the masked token.
\textbf{Prefix style} aims to let LMs generate an answer given a prefix, which means the entire input comes before the final prediction. For example, the input \textit{"I love this movie"} will be formulated into \textit{"I love this movie. Is this review positive or negative?"},  and the model will be asked to generate the answer.

\subsection{Prompt Position}
\label{Setup: prompt position}
Prompt position is the variable of interest in our study. We take into account the position where prompt tokens can be inserted and enumerate a broad range of permutations to test. 

\begin{figure}[!htbp]
  \centering
  \begin{subfigure}{0.48\textwidth}
    \centering
\includegraphics[width=0.65\linewidth]{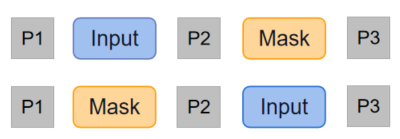}
    \caption{single-sentence tasks}
    \label{fig:cloze_single}
  \end{subfigure}
  \hspace{1cm}
  \begin{subfigure}{0.48\textwidth}
    \centering
    \includegraphics[width=\linewidth]{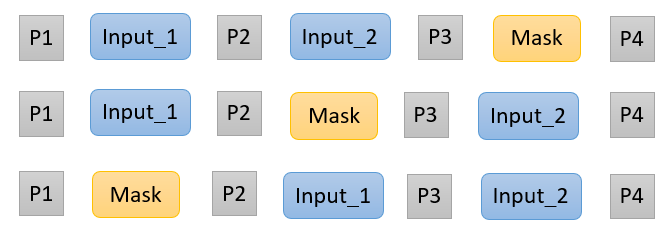}
    \caption{sentence-pair tasks}
    \label{fig:cloze_pair}
  \end{subfigure}
  \caption{Insertion positions for cloze-style prompts}
  \label{fig:cloze_style_position}
\end{figure}

Concretely, for \textbf{Cloze style} prompts, as shown in Figure \ref{fig:cloze_style_position}, we consider the relative position of the [mask] token to the input. There are $m$ types of Input-[Mask] concatenations ($m=2$ for single-sentence tasks and $m=3$ for sentence-pair tasks\footnote{Regarding the sentence-pair classification task, we maintain the expected task input sequence order to narrow our focus on the prompt position (e.g. premise then hypothesis).}), each with $n$ potential locations that could insert prompt sequences ($n=3$ for single-sentence tasks and $n=4$ for sentence-pair tasks). In contrast to \citet{wu2022idpg} who inserts a single sequence of prompt tokens at different positions, we insert at least one and at most $n$ prompt series per concatenation, yielding a total of $m\cdot(2^n-1)$ prompt positions. For \textbf{Prefix style} prompts, we explore $n$ insertion points ($n=2$ and $n=3$ corresponding to the first row in subfigure \ref{fig:cloze_single} and subfigure \ref{fig:cloze_pair}) without considering the [mask] token ($m=1$), which results in $2^n-1$ different prompt positions.

\section{Effect on Gradient-based Prompting}
\label{gradient_based_experiments}

\subsection{Setup}
\label{gradient_based_setup}
For gradient-based approaches, both discrete \footnote{We do not take the automated search approach in discrete space, as this may result in different vocabulary in templates and potentially obfuscate the impact prompt position has in our results.} and continuous methods are investigated. To focus on studying the effect of the prompt position itself, we implement two vanilla approaches: 

\textbf{Prompt-based fine-tuning:} For discrete prompt, we fine-tune all the LM’s parameters with the input restructured within a manual prompt template as per \cite{schick2020IPET,gao2020LMBFF}. 

\textbf{Prompt tuning:} For continuous prompt, we instantiate standard prompt tuning \cite{lester2021power}, which only tunes the continuous prompts tokens prepended to the input layer with the language model frozen. Besides, we incorporate both cloze and prefix styles, leading to four types of prompts for empirical investigation: \textit{\textbf{cloze manual prompt}}, \textit{\textbf{cloze continuous prompt}}, \textit{\textbf{prefix manual prompt}} and \textit{\textbf{prefix continuous prompt}}.

\paragraph{Models:} We choose language models which are popular in the NLU research literature. As per 
 \citet{gao2020LMBFF}, we use RoBERTa-large \cite{roberta} to predict the masked token based on the cloze-style prompt. To generate answers from prefix-style prompts, we use T5-Large language model adaption as per \citet{lester2021power}, which is pre-trained for 10K steps with language modelling objectives without mixing downstream tasks. We additionally experiment with T5-XL (3B) on prefix continuous prompt to investigate the relatively larger model. All our gradient-based experiments are conducted using the OpenPrompt framework\footnote{https://github.com/thunlp/OpenPrompt}\cite{ding2021openprompt}. 

\paragraph{Datasets:}
We examine the above approaches on five commonly used natural language understanding datasets as per \cite{gao2020LMBFF,lester2021power}. The datasets span various tasks: sentiment analysis (CR \cite{CRhu2004mining}, SST-2 \cite{wang2018glue}), question classification (TREC \cite{TREC}), question answering (BoolQ \cite{wang2019superglue}), and natural language inference (RTE \cite{wang2019superglue}), broadly classified into single-sentence (SST-2, CR, TREC) and sentence-pair categories (RTE, BoolQ). For datasets from GLUE and SuperGLUE, we use the original development sets for testing, and for the rest, we follow the testing sets as per \citet{gao2020LMBFF}. See Table \ref{tab:dataset} in Appendix for details.

We measure the effect of the prompt position by the model’s few-shot performance. We construct $D_{train}$ and $D_{dev}$ with K samples per label from the original training data, with K ranging from 16 to 128. We calculate the average accuracy across five randomly sampled $D_{train}$ and $D_{dev}$ splits, using the same fixed set of seeds as per \citet{gao2020LMBFF}.

\begin{table*}[ht!]
 \centering
\resizebox{0.9\textwidth}{!}{
\begin{tabular}{ll|ll|ll|ll|ll|ll}
  \hline\noalign{\smallskip}
        &       & SST-2 &              & CR   &             & TREC &             & RTE  &             & BoolQ &             \\
K-size &
  Method &
  Var &
  Best (\(\Delta\)) &
  Var &
  Best (\(\Delta\)) &
  Var &
  Best (\(\Delta\)) &
  Var &
  Best (\(\Delta\)) &
  Var &
  Best (\(\Delta\)) 
   \\
     \hline\noalign{\smallskip}
16  & CM    & 5.1   & 90.8         & 3.3  & 91.2        & 3.3  & 85.7 (+2.9) & 17.8 & 71.2 (+3.7) & 14.8  & 69.5 (+2.0)  \\
        & CC    & 23.1  & 86.3 (+0.2)  & 18.7 & 84.6 (+0.8) & 17.4 & 65.1 (+6.8) & 10.2 & 57.0 (+1.4) & 17.1  & 61.1 (+1.0)   \\
        & PM    & 3.8   & 89.5         & 2.7  & 92.7 (+0.2) & 0.8  & 85.0 (+0.3) & 6.2  & 61.7 (+3.1) & 6.7   & 64.2 (+3.0)   \\
        & PC    & 13.7  & 70.6 (+13.4) & 26.4 & 86.8        & 3.9  & 71.8 (+2.4) & 4.4  & 52.1        & 4.3   & 49.6 (+4.3)  \\
        & PC-XL & 12    & 84.9         & 18.3 & 86.7        & 5.5  & 80.8 (+5.5) & 6.6  & 53.1 (+1.7) & 6.1  & 51.8 (+4.8)   \\
          \hline\noalign{\smallskip}
32  & CM    & 4.4   & 91.4 (+0.7)  & 2.5  & 92.0 (+0.9) & 1.6  & 88.8 (+1.3) & 20.6 & 74.7 (+1.8) & 17    & 72.1 (+3.8)   \\
        & CC    & 17.9  & 89.7         & 14.8 & 89.2 (+0.2) & 12.9 & 74.8 (+1.7) & 10.5 & 59.6 (+0.5) & 14.9  & 63.5 (+6.1)  \\
        & PM    & 0.9   & 92           & 1.9  & 93.0 (+0.1) & 0.4  & 88.3 (+0.4) & 11.4 & 66.9 (+7.0) & 6.3   & 71.1         \\
        & PC    & 12.8  & 75.5 (+8.4)  & 17   & 89.4        & 5.6  & 80.4 (+5.6) & 2.5  & 52.6 (+0.2) & 7.9   & 57.3 (+5.2)  \\
        & PC-XL & 4.1   & 90.6         & 2.1  & 90.4 (+0.4) & 2.8  & 84.9 (+2.8) & 5.3  & 54.4        & 4.8   & 53.7 (+4.8)  \\
          \hline\noalign{\smallskip}
64  & CM    & 2.4   & 92.4 (+0.9)  & 1.6  & 92.3 (+0.2) & 1.3  & 92.3 (+1.1) & 22.9 & 77.3 (+4.3) & 14    & 73.8 (+0.6)  \\
        & CC    & 10.9  & 91.4 (+1.2)  & 7.8  & 91.1 (+1.3) & 10.1 & 81.8        & 14.4 & 63.5 (+4.3) & 11.4  & 62.6 (+0.5)  \\
        & PM    & 0.7   & 93.1 (+0.7)  & 0.6  & 93.7        & 0.4  & 90.5        & 12.9 & 72.9 (+4.8) & 3.1   & 73.8         \\
        & PC    & 14    & 89.1 (+5.3)  & 13.3 & 92          & 3.5  & 85.5 (+3.5) & 6.8  & 57.3        & 2.8   & 52.9 (+0.6)           \\
        & PC-XL & 1.3   & 92.8         & 0.8  & 92.7 (+0.8) & 1.4  & 88.6        & 10.4 & 62.3        & 5.8   & 55.1 (+2.5)   \\
          \hline\noalign{\smallskip}
128 & CM    & 1.7   & 93           & 1.2  & 92.6        & 1.1  & 94.6 (+0.7) & 18.1 & 79.8 (+1.0) & 13.3  & 75.7         \\
        & CC    & 7.6   & 92.6 (+2.3)  & 4.8  & 92.1 (+1.4) & 6.2  & 88.1 (+0.7) & 12.2 & 63.1 (+3.0) & 13.9  & 65.1 (+4.7)  \\
        & PM    & 0.6   & 93.4 (+0.2)  & 0.4  & 93.9 (+0.2) & 0.6  & 92.2 (+0.6) & 9    & 75.2 (+3.8) & 2.7   & 76.5 (+1.0)  \\
        & PC    & 7.7   & 88.5         & 5.7  & 91.6        & 1.6  & 89.4 (+1.6) & 9.6  & 61.2 (+2.1) & 4.2   & 54.9 (+4.2)  \\
        & PC-XL & 1.4   & 93.3         & 0.5  & 93.9        & 1    & 91.7 (+0.5) & 9.9  & 64.4        & 5.4  & 58.2 (+5.4)  \\
          \hline\noalign{\smallskip}
\end{tabular}
}
\caption{The effect of prompt position in few-shot gradient-based prompting across K training samples per label.  'Var' indicates the performance disparity between the least and most effective positions. 'Best' is the accuracy at the optimal position. (\(\Delta\)) is the accuracy delta of the optimal position compared to the reference position. If the optimal position is the reference position, then (\(\Delta\)) is not reported. We use the following abbreviations. \textbf{CM}: cloze manual prompt; \textbf{CC}: cloze continuous prompt; \textbf{PM}: prefix manual prompt; \textbf{PC}: prefix continuous prompt; \textbf{PC-XL}: prefix continuous prompt with the T5-XL model. See Appendix \ref{appendix: all results} for the full results.}
\label{tab:sum}
\end{table*}

\begin{table}[htbp!]
\centering
\resizebox{0.49\textwidth}{!}{
\begin{tabular}{l}
\hline
Prefix manual prompt  \\
\hline
Question: Is this sentence positive or negative?  Answer: \{text\}  \\
\textit{\{text\} Question: Is this sentence positive or negative?  Answer:} \\
Question: Is this sentence positive or negative? \{text\} Answer:  \\
\hline
\end{tabular}
}
\caption{Example templates with different positions for SST-2. The italic row indicates the default template position in \citet{evalharness}.}
\label{tab:position-example}
\end{table}

\paragraph{Positions:}
As explained in Section \ref{Setup: prompt position}, we experiment with distinct prompt positions tailored to different prompt styles: there are $14$ positions for cloze-style prompts in single-sentence tasks and $45$ positions in sentence-pair tasks; for prefix-style prompts, we use $3$ positions for single-sentence tasks and $7$ for sentence-pair tasks.

To mitigate the influence of vocabulary in the manual prompt, we employ a reference template from prior research and only change the position where these words are inserted (See Table \ref{tab:position-example} for simple examples). In practice, there would be different situations for each prompt position, particularly with discrete prompts. For example, when inserting task input between two prompt sequences, the choice of which sequence to place at the beginning or end can vary.  We prioritize templates that maintain grammatical coherence, selecting one template per prompt position\footnote{All prompts obtained by altering the position presented in this paper were chosen before evaluation to avoid any selection bias.}. With regards to continuous prompts, when inserting multi-prompt series, we simply separate continuous tokens used in the single prompt sequence equally to mitigate the effect of prompt length. All templates and verbalizers we used are described in Appendix \ref{appendix: reference prompt}, along with their respective prompt position options for different tasks (Appendix \ref{appendix: all results}).

\begin{table*}[ht]
  \centering
  \large
\resizebox{0.9\textwidth}{!}{
    \begin{tabular}{llll}
\hline\noalign{\smallskip}
      K-size & Method & RTE & BoolQ \\
      \hline\noalign{\smallskip}
      K=16 & CM & the Answer: \{text\_a\} . {[}mask{]} \{text\_b\} Question: ? &  Answer: \{text\_a\} . Question: \{text\_b\} ? {[}mask{]} . \\
      & CC & \textit{\textbf{P}} \{text\_a\} \textit{\textbf{P}} {[}mask{]} \{text\_b\}  \textit{\textbf{P}} 
      & \textit{\textbf{P}} \{text\_a\} \textit{\textbf{P}} {[}mask{]} \{text\_b\} \\
      & PM & \{text\_a\} Question: True or False? Answer: \{text\_b\} & Question: \{text\_a\} \{text\_b\} ? Answer: \\
      & PC & \textit{\textbf{P}} \{text\_a\} \{text\_b\} 
      & \textit{\textbf{P}} \{text\_a\} \textit{\textbf{P}} \{text\_b\}  \\
       & PC-XL & \{text\_a\} \textit{\textbf{P}} \{text\_b\} \textit{\textbf{P}}
      & \{text\_a\} \textit{\textbf{P}} \{text\_b\} 

\\
      \hline\noalign{\smallskip}
      K=32 & CM & the Answer: \{text\_a\} . {[}mask{]} \{text\_b\} Question: ? & \{text\_a\} . Question: ? Answer: . {[}mask{]} \{text\_b\} \\
      & CC & \textit{\textbf{P}} \{text\_a\} {[}mask{]} \{text\_b\} \textit{\textbf{P}} & \{text\_a\} {[}mask{]} \{text\_b\}  \textit{\textbf{P}} \\
      & PM & Question: True or False? Answer: \{text\_a\} \{text\_b\} & \{text\_a\} Question: \{text\_b\} ? Answer: \\
      & PC & \textit{\textbf{P}} \{text\_a\} \textit{\textbf{P}} \{text\_b\} 
      & \textit{\textbf{P}} \{text\_a\} \textit{\textbf{P}} \{text\_b\} 
\\
       & PC-XL & \textit{\textbf{P}} \{text\_a\} \{text\_b\} 
      & \{text\_a\} \textit{\textbf{P}} \{text\_b\} 
      \\
      \hline\noalign{\smallskip}
      K=64 & CM & the Answer: \{text\_a\} . {[}mask{]} \{text\_b\} Question: ? & \{text\_a\} . Question: ? \{text\_b\} {[}mask{]} Answer: . \\
      & CC & \{text\_a\} {[}mask{]} \{text\_b\}  \textit{\textbf{P}} & \{text\_a\} {[}mask{]} \{text\_b\} \textit{\textbf{P}} \\
      & PM & Answer: \{text\_a\} Question: True or False? \{text\_b\} & \{text\_a\} Question: \{text\_b\} ? Answer: \\
      & PC & \textit{\textbf{P}} \{text\_a\} \{text\_b\} 
      & \{text\_a\} \{text\_b\} \textit{\textbf{P}} 
\\
       & PC-XL & \textit{\textbf{P}} \{text\_a\} \{text\_b\} 
      & \textit{\textbf{P}} \{text\_a\} \textit{\textbf{P}} \{text\_b\}  
      \\
      \hline\noalign{\smallskip}
      K=128 & CM & the Answer: . \{text\_a\} Question: ? \{text\_b\} {[}mask{]} & \{text\_a\} . Question: \{text\_b\} ? Answer: {[}mask{]} . \\
      & CC & \textit{\textbf{P}}                  \{text\_a\} {[}mask{]} \{text\_b\}  \textit{\textbf{P}} 
      & \{text\_a\} {[}mask{]} \{text\_b\}  \textit{\textbf{P}} \\
      & PM & True or False? \{text\_a\} Question: \{text\_b\} Answer: & \{text\_a\} Question: ? Answer: \{text\_b\} \\
      & PC & \{text\_a\} \{text\_b\}  \textit{\textbf{P}} 
      & \textit{\textbf{P}} \{text\_a\} \textit{\textbf{P}} \{text\_b\}  \textit{\textbf{P}} \\
       & PC-XL & \textit{\textbf{P}} \{text\_a\} \{text\_b\} 
      & \textit{\textbf{P}} \{text\_a\} \textit{\textbf{P}} \{text\_b\} \textit{\textbf{P}} \\
      \hline\noalign{\smallskip}
    \end{tabular}
  }
  \caption{The optimal prompt positions on RTE and BoolQ. \textit{\textbf{P}} denotes continuous prompt tokens.}
  \label{tab:best-pair}
\end{table*}

\subsection{Results}
As demonstrated in Table \ref{tab:sum}, for single-sentence tasks - SST-2, CR and TREC, the influence of prompt position is relatively small when using manual prompts, while significant performance variations arise when continuous prompts are employed. With the K-size increases, the differences between all methods tend to diminish. We note that the optimal prompt position may not always align with the reference position across different methods especially in the TREC dataset. Yet, the accuracy gap compared to the reference position also becomes smaller when K is set to 128.

For sentence-pair tasks, RTE and BoolQ, a substantial performance variation is observed across all methods. As K increases, the variance between different prompt positions persists except for the case of the prefix manual prompt in BoolQ. Similar to single-sentence tasks, the reference prompt position does not consistently produce optimal results across all methods. Notably, even when the K is set to 128, there are instances where a noticeable difference exists between the best-performing prompt position and the reference position. For example, in RTE, the prefix manual prompt shows a 3.8 percentage point difference in performance, and in BoolQ, the prefix continuous prompt has a 5.4 percentage point difference (Table \ref{tab:sum}).

We additionally experiment with T5-3B on prefix continuous prompt, more commonly used in prompt tuning methods, to investigate the effect of prompt position in a relatively large model.  Our results indicate that a larger scale helps to reduce the performance difference in single-sentence tasks, especially when $K>16$. However, the variance remains considerable in sentence-pair tasks.

In general, sentence-pair tasks are more susceptible to the influence of prompt position compared to single-sentence tasks, whereas continuous prompts exhibit higher sensitivity to position compared to manual prompts. 

\subsection{Discussion} \label{gradient-based-discussion}
In our main paper, we detail the optimal prompt positions for sentence-pair tasks in Table \ref{tab:best-pair}, while those for single-sentence tasks are in Table \ref{tab:best-single} of the Appendix. Table \ref{tab:best-pair} shows that the optimal prompt position is not consistently shared across different datasets when employing the same prompt method. For example, for the prefix continuous prompting (PC), the RTE task often prefers prompt tokens inserted at the start, especially with the T5-3B model. In contrast, BoolQ shows a preference for the position "\textit{\textbf{P}} \{text\_a\} \textit{\textbf{P}} \{text\_b\} \textit{\textbf{P}}" in both T5-Large and T5-XL models (K=128). Also, we have noticed that the optimal position varies depending on K size, indicating that the distribution of input samples holds an influence. Besides, there is no clear superiority between inserting multiple prompt sequences and a single prompt sequence. However, the relative position of the [mask] token in cloze-style prompts indeed affects the model performance, which is consistent with the findings of \citet{gao2020LMBFF}. Typically, placing the [mask] token between the two inputs is often favoured in RTE. We conduct supplementary experiments with null templates which will be further discussed in Appendix  \ref{appendix:mask_token}.

It is worth noting that grammar doesn't always dictate the performance of manual prompts. This can be observed where grammatically incorrect prompts often achieve the best performance, and the performance difference between grammatically correct and incorrect prompts is not always negligible. For example, in the BoolQ dataset within a cloze-style manual prompt (K = 32), \textit{"\{text\_a\} . Question: ? Answer: . {[}mask{]} \{text\_b\}"} (shown in Table \ref{tab:best-pair}) outperforms the reference prompt \textit{"\{text\_a\} . Question: \{text\_b\} ? Answer: {[}mask{]} ."} \cite{IPET2} by 3.84 percentage points (Table \ref{tab:sum}). This phenomenon suggests that prompts considered reasonable by humans may not necessarily be effective for language models, which is consistent with \citet{P-tuning}. It implies that factors beyond grammar contribute to performance outcomes.

\begin{table*}[htbp!]
\large
\centering
\renewcommand{\arraystretch}{1.2}  \resizebox{\textwidth}{!}{
    \begin{tabular}{l|ll|ll|ll|ll|ll|ll}
    \hline\noalign{\smallskip}
     & \multicolumn{2}{l}{T5-3B} & \multicolumn{2}{l}{Flan-T5-3B} & \multicolumn{2}{l}{T5-11B} & \multicolumn{2}{l}{Flan-T5-11B} & \multicolumn{2}{l}{LLaMA-13B} & \multicolumn{2}{l}{Flan-LLaMA-13B} \\
     & Var   & Best (\(\Delta\)) & Var   & Best (\(\Delta\)) & Var   & Best (\(\Delta\)) & Var   & Best (\(\Delta\)) & Var   & Best (\(\Delta\)) & Var   & Best (\(\Delta\)) \\
    \hline\noalign{\smallskip}
Causal Judgement         & 4.3   & 48.7        & 3.2     & 63.1        & 1.6    & 50.3 (+1.6) & 3.2      & 60.4 (+0.5)  & 3.2
   & 50.8 (+3.2)       & 1.6        & 59.4  \\
Disambiguation QA        & 1.6   & 33.2        & 2.0 &	68.4 (+2.0) & 5.2    & 34.8        & 2.8	& 68.4 (+1.2)  & 26.0   & 61.6        & 6.0         & 63.2
         \\
Sports Understanding     & 2.0   & 48.0        & 4.0	& 58.8
 & 2.8    & 51.6 (+2.8) & 0.4	& 69.2         & 5.6   & 68.4 (+4.0) & 2.8      & 67.2         \\
Navigate                 & 14.0    & 42.0          & 5.6	& 60.4      & 7.2    & 47.6 (+2.8) & 2.8      & 61.6 (+2.8)  & 0     & 58.0          & 1.2        & 58.0  \\
Logical Deduction (5)    & 0     & 22.4        & 8.0	& 50.4 (+3.6) & 4.8    & 23.2 (+3.2) & 5.2	& 55.2
  & 8.0   & 26.0 & 10.0       & 37.6 (+0.4)         \\
Logical Deduction (7)    & 1.6   & 15.2 (+1.6) & 6.0 & 52.4
       & 1.2    & 18.8 (+0.4) & 3.6 & 59.2 (+0.4) & 2.8  & 18.8 & 11.2      & 35.6 (+6.4)  \\
Logical Deduction (3)    & 36.4  & 36.4        & 6.4 & 64.4 (+1.6) & 1.6    & 36.0          & 2 & 72.4 (+2.0)           & 6.8   & 41.6        & 1.2        & 43.2 (+1.2)  \\
Penguins in a Table      & 8.2   & 24.0 (+8.2) & 5.5 & 37.7
       & 4.1    & 25.3 (+3.4) & 9.6      & 44.5 (+2.0)  & 1.4    & 26.7        & 9.6       & 37.7           \\
Salient Translation Err. & 4.0     & 16.0 (+4.0)   & 11.2    & 45.2        & 0      & 12.0          & 6.8      & 51.2         & 14.8  & 38.4 & 14.4       & 36.0 (+1.2)  \\
Movie Recommendation     & 2.0   & 27.2 (+2.0) & 20.8    & 65.2 (+10.0)  & 15.6   & 44.0          & 32.8     & 71.6 (+11.6) & 34.0  & 80.0 (+8.8)
   & 15.6      & 80.0 (+11.6) \\
\hline\noalign{\smallskip}
All task (avg)           & 7.41  &             & 7.27    &             & 4.41   &             & 6.92     &              & 10.26  &             & 7.36       &         \\
\hline\noalign{\smallskip}
\end{tabular}
}
\caption{The effect of prompt position in few-shot direct prompting on BBH benchmark.}
\label{tab:bbh-few-direct}
\end{table*}

\begin{table*}[htbp!]
\large
\centering
\renewcommand{\arraystretch}{1.2}
\resizebox{\textwidth}{!}{
    \begin{tabular}{l|ll|ll|ll|ll|ll|ll}
    \hline\noalign{\smallskip}
     & \multicolumn{2}{l}{T5-3B} & \multicolumn{2}{l}{Flan-T5-3B} & \multicolumn{2}{l}{T5-11B} & \multicolumn{2}{l}{Flan-T5-11B} & \multicolumn{2}{l}{LLaMA-13B} & \multicolumn{2}{l}{Flan-LLaMA-13B} \\
     & Var   & Best (\(\Delta\)) & Var   & Best (\(\Delta\)) & Var   & Best (\(\Delta\)) & Var   & Best (\(\Delta\)) & Var   & Best (\(\Delta\)) & Var   & Best (\(\Delta\)) \\
    \hline\noalign{\smallskip}
Causal Judgement &
  0.5 &
  43.3 &
  5.9 &
  60.4 (+3.2) &
  2.6 &
  48.1 &
  0.5 &
  57.2 (+0.5) &
  2.7 &
  54 &
  5.4 &
  57.8 (+2.7) \\
Disambiguation QA &
  2.4 &
  28.4 &
  1.6 &
  69.2 &
  2 &
  30.0 (+0.8) &
  1.2 &
  64.8 (+0.8) &
  8.4 &
  50.0 (+8.4) &
  1.6 &
  39.2 (+1.6) \\
Sports Understanding &
  1.2 &
  46.8 (+0.8) &
  5.2 &
  58.0 (+2.0) &
  4 &
  59.2 (+4.0) &
  4.4 &
  64.0 (+2.8) &
  1.6 &
  78.4 (+1.6) &
  2.8 &
  77.2 \\
Navigate &
  5.2 &
  27.6 (+0.8) &
  6.8 &
  56.8 (+0.4) &
  4.4 &
  32.8 &
  4 &
  61.6 &
  2.8 &
  59.6 (+0.8) &
  2.4 &
  57.6 \\
Logical Deduction (5) &
  4.0 &
  20.8 (+4.0) &
  10 &
  37.6 &
  8.4 &
  21.6 (+1.2) &
  9.2 &
  51.2 &
  1.2 &
  29.6 &
  4.4 &
  32.8 (+4.4) \\
Logical Deduction (7) &
  3.2 &
  14.4 (+1.6) &
  11.6 &
  30 &
  6.4 &
  17.2 &
  18.8 &
  53.6 &
  3.2 &
  22.8 (+1.2) &
  0.8 &
  20.0 (+0.4) \\
Logical Deduction (3) &
  2.8 &
  38.8 (+2.8) &
  2.8 &
  55.6 &
  8.4 &
  40.4 &
  11.2 &
  67.2 &
  3.6 &
  49.2 &
  2.4 &
  57.6 (+2.0) \\
Penguins in a Table &
  4.2 &
  29.5 (+4.2) &
  2.7 &
  26.7 (+1.4) &
  6.9 &
  24 &
  13.7 &
  43.8 (+6.8) &
  5.5 &
  40.4 &
  3.5 &
  42.5 \\
Salient Translation Err. &
  2.4 &
  9.2 &
  7.6 &
  25.6 (+2.0) &
  8 &
  14 &
  27.2 &
  38.4 &
  5.2 &
  20.4 (+5.2) &
  0 &
  30.8 \\
Movie Recommendation &
  3.6 &
  26 &
  15.6 &
  53.2 &
  18.4 &
  47.2 &
  11.2 &
  44.4 &
  15.2 &
  70 (+15.2) &
  25.6 &
  74 \\
  \hline\noalign{\smallskip}
  All task (avg) &
  2.95
 &
   &
  6.98
 &
   &
  6.95
 &
  &
  10.14
 &
   &
  4.94
 &
   &
  4.89
 &
   \\
  \hline\noalign{\smallskip}
\end{tabular}
}
\caption{The effect of prompt position in few-shot CoT prompting on BBH benchmark.}
\label{tab:bbh-few-cot}
\end{table*}

\section{Effect on Gradient-free Prompting }\label{gradient_free_experiments}
\subsection{Setup}\label{gradient_free_setup}
We explore classic gradient-free prompting methods within both zero-shot and few-shot paradigms. For few-shot settings, in-context learning is investigated via \textbf{direct prompting} \cite{GPT3} as well as \textbf{chain-of-thought} (CoT) prompting,  where models provide a reasoning step prior to the final response \cite{wei2022chain}. For zero-shot settings, we only consider the direct prompting. 

We examine the in-context learning performance of SST-2 using a 16-shot setting (K=8), which is the maximum length accommodated within 512 tokens, on the T5-Large model. We find a consistent trend that the position variance is significant and the reference position is sub-optimal. However, the overall performance of in-context learning significantly lags behind that of prompt-based fine-tuning by employing the same manual prompt (see details in Appendix \ref{appendix: SST-2}). We shift our focus to relatively larger models in this section.
\paragraph{Models:}
We investigate two T5-series models, T5-3B and T5-11B \cite{raffel2023exploring}, and a decoder-only model LLaMA-13B \cite{touvron2023llama}, which has sufficient capability to assess the impact of prompt position without any fine-tuning. Additionally, we conduct experiments with their instruction-tuned variants, Flan-T5-XXL \cite{chung2022scaling} and Flan-LLaMA \cite{flanllama}. These variants have been pre-trained on a diverse set of data sources utilizing an array of instruction template types that incorporate a wide spectrum of vocabulary and positional variations.

For evaluation, we measure accuracy using the exact match in the few-shot direct prompting as per \citet{suzgun2022bbh,chung2022scaling}. Within the CoT setup, we extract the first word after the phrase \textit{‘[T|t]he answer is’}, or capture the full response if no such pattern is present. For the zero-shot setting, we compute the likelihood of each target option and select the option with the highest log-likelihood as per \citet{GPT3,sanh2022multitask,wei2022instruction}. 

\begin{table*}[ht!]
 \centering
\resizebox{0.85\textwidth}{!}{
    \begin{tabular}{l|ll|ll|ll|ll}
    \hline\noalign{\smallskip}
      & \multicolumn{2}{l}{Flan-T5-3B} & \multicolumn{2}{l}{Flan-T5-11B} & \multicolumn{2}{l}{LLaMA-13B} & \multicolumn{2}{l}{Flan-LLaMA-13B} \\
     & Var   & Best (\(\Delta\)) & Var   & Best (\(\Delta\)) & Var   & Best (\(\Delta\)) & Var   & Best (\(\Delta\)) \\
    \hline\noalign{\smallskip}
Causal Judgement                      & 1.1     & 61           & 1.6      & 61.5 (+0.5) & 2.7   & 53.5 (+1.6) & 5.3        & 58.8 (+3.2) \\
Disambiguation QA                     & 2.4     & 69.2 (+2.4)  & 4        & 68.8        & 20.8  & 52.8        & 16.4       & 60          \\
Sports Understanding                  & 2.4     & 53.6 (+1.6)  & 2.4      & 61.6 (+2.4) & 8.8   & 60.8        & 5.6        & 64.4        \\
Navigate                               & 3.2     & 60.8 (+2.4)  & 1.6      & 60.8        & 7.6   & 49.6 (+7.6) & 1.2        & 59.2 (+1.2) \\
Logical Deduction (5)      & 4       & 50.4 (+0.4)  & 7.2      & 56.8 (+1.6) & 3.2   & 21.2        & 15.2       & 38          \\
Logical Deduction (7)     & 6       & 53.2 (+0.8)  & 8        & 62.4 (+1.2) & 0     & 14.8        & 15.2       & 33.6        \\
Logical Deduction (3)     & 4.4     & 64.8         & 5.6      & 75.2        & 7.6   & 39.6 (+3.2) & 7.2        & 50 (+0.4)   \\
Penguins in a Table                 & 8.9     & 41.1         & 9.5      & 47.9 (+6.1) & 10.2  & 26          & 8.9        & 31.5        \\
Salient Translation Err & 3.2     & 42.4         & 0        & 50          & 4.4   & 18 (+4.0)   & 5.2        & 24.8        \\
Movie Recommendation                  & 28.8    & 67.2 (+13.2) & 14.8     & 55.2 (+8.8) & 34.4  & 56          & 38         & 67.6 (+2.0) \\
\hline\noalign{\smallskip}
All task (avg)                         & 6.44    &              & 5.47     &             & 9.97  &             & 11.82      &            \\
\hline\noalign{\smallskip}
\end{tabular}
}
\caption{The effect of prompt position in zero-shot direct prompting on BBH benchmark.}
\label{tab:bbh-zero}
\end{table*}

\paragraph{Datasets:}
We evaluate the sub-tasks of BIG-Bench Hard (BBH) \cite{suzgun2022bbh}, a challenging benchmark from BIG-Bench\footnote{We choose tasks with defined task specifications from \citet{srivastava2023beyond} to ensure that we can effectively alter prompt positions.}, for the fact that instruction-tuned models were not exposed to it during training. The tasks involve not only NLU (e.g. Disambiguation QA) but also reasoning (e.g. Navigate, Logical Deduction), the Use of World Knowledge (e.g. Sports Understanding, Movie Recommendation, Causal Judgement), Table Parsing (e.g. Penguins in a Table) and the Multilingual task (e.g. Salient Translation Error detection). Following \citet{suzgun2022bbh}, we employ the officially provided prompts, each accompanied by three few-shot examplers, for both chain-of-thought and direct prompting.

\paragraph{Positions:}
This experiment is constrained to one prompt type, prefix manual prompt, a choice informed by the nature of the models and methods we employed here. Given the only one task-input in the BBH benchmark, we investigate three prompt positions for each sub-task: insertion at the front, the rear, or on both sides of the input. We play the relative position between input and prompt within the exemplar delimiters (e.g., "Q:"/"A:")\footnote{Our rationale for this choice is that templates with exemplar delimiters consistently yield better results and also align with the approach of \citet{chung2022scaling} in the creation of few-shot templates for instruction fine-tuning.}. All templates and their variants of positions are detailed in Appendix \ref{appendix: reference prompt}.

\subsection{Results}
\paragraph{Few-shot:} As illustrated in Table \ref{tab:bbh-few-direct} and Table \ref{tab:bbh-few-cot}, both direct prompting and CoT prompting exhibit significant performance variances across different prompt positions in most scenarios. Also, the optimal position does not always align with the default settings. Notably, the performance disparity between the default (used in \citet{suzgun2022bbh}) and the best outcome (obtained by altering the position) is significant sometimes. For instance, in the movie recommendation, there is an 11.6 per cent performance gap for direct prompting on Flan-LLaMA-13B, and a 15.2 per cent gap for CoT prompting on LLaMA-13B (changing from "\textit{Find a movie similar to \{Input\} \textbackslash n\{Options}\}" to "\textit{\{Options\} \textbackslash n Find a movie similar to \{Input\}}").

When comparing larger model scales (3B VS 11B), the variance persists, and in some cases, it even increases. For instance, in the task of three-object logical deduction, the variance among different positions rises from 2.8 per cent to 11.2 per cent on Flan T5-based models using the chain-of-thought method (Table \ref{tab:bbh-few-cot}). Similarly, and quite surprisingly, we observe that instruction tuning does not necessarily reduce the performance difference between positions; in fact, it can sometimes exacerbate them.
\paragraph{Zero-shot:} In zero-shot scenarios, T5 plain models show only marginal improvements or even worse than random, so our attention is primarily on the other four models. As indicated in Table \ref{tab:bbh-zero}, we observe similar trends to those in the few-shot setting. There are noticeable performance differences between various prompt positions in most cases, and the position yielding the best performance does not consistently match the reference position.

\begin{table*}[htbp!]
\centering
\renewcommand{\arraystretch}{1.2}
  \resizebox{0.85\textwidth}{!}{
\begin{tabular}{l|lll|lll|lll|lll}
\hline\noalign{\smallskip}
                      & \multicolumn{3}{c}{Flan-T5-3B} & \multicolumn{3}{c}{ Flan-T5-11B} &       \multicolumn{3}{c}{LLaMA-13B} & \multicolumn{3}{c}{Flan-LLaMA-13B} \\
                      & Direct & CoT        & 0-shot & Direct & CoT         & 0-shot & Direct & CoT       & 0-shot & Direct & CoT            & 0-shot \\
                      \hline\noalign{\smallskip}
Causal Judgement         & B  & R & B/R & R & F/R  & R & F & B  & F & B  & F      & F \\
Disambiguation QA        & F & B & F     & F      & F & B & B & F & B  & B  & F/R & B  \\
Sports Understanding  & F   & R       & B   & F/R  & R        & R   & B   & B      & F  & F  & F          & F  \\
Navigate              & B   & F      & R   & R   & B        & B   & -      & F     & R   & B  & B           & R   \\
Logical Deduction (5) & R   & B       & R   & B/R   & B        & R   & B   & B/R & B   & R   & R           & B   \\
Logical Deduction (7) & B   & B       & R   & R   & B        & R   & B   & F     & -      & R   & F          & B   \\
Logical Deduction (3)    & R  & B & B/R & F & B  & B & B & B  & R  & R & R       & R  \\
Penguins in a Table   & B   & F      & B   & F  & F       & F  & B     & B      & B   & B   & B           & B   \\
Salient Translation Err. & B  & R & B      & F/B & B  & -    & B & F & F & F & -          & B  \\
Movie Recommendation  & R   & B       & R   & R   & B        & R   & R   & F     & B   & F  & B           & R  
\\\hline\noalign{\smallskip}
    \end{tabular}
  }
  \caption{The optimal prompt positions of different models on BBH. 'F' for Front Insertion, 'B' for Both Sides and 'R' for Rear Placement.}
  \label{tab:best-position-bbh}
\end{table*}

\subsection{Discussion}
In Table \ref{tab:best-position-bbh}, we present the optimal prompt positions for various models across different tasks. It reveals that there is no universal best position across tasks, models and methods,  echoing our findings discussed in Section \ref{gradient-based-discussion}. While the default prompt position is typically set to "both" (namely prompt sequences are inserted both in the front and rear of the input), this doesn't yield the best performance all the time. Besides, the optimal position in the zero-shot setting may not correspond to the one in the few-shot setting. This is nuanced and consistent with our discussion in Section \ref{gradient-based-discussion} that the input distribution may hold influence. 

The grammar tends to better-preserved compared to the cases setup in Section \ref{gradient_based_experiments} when we alter positions here. Despite this, there is still a significant variance in performance between different positions. We suspect this variance may stem from the "favourite" position of the language model, influenced by similar formatted data encountered during pre-training or instruction-tuning. As previously noted, instruction-tuned models do not necessarily diminish this variance, despite typically incorporating a wide range of vocabulary and positional variations in their training templates. We have reviewed the ten templates for each dataset in \citet{chung2022scaling} and found an uneven distribution of position variety. For instance, in the SST-2 task, a majority of their instruction tuning templates position the task input between the task specification and options. In contrast, in the CNN Daily Mail summarization task \cite{CNN-DM-see-etal-2017-get}, their templates often place the task input at the beginning. 

To further explore the potential reason for this variance, we conduct a preliminary experiment on SST-2 using Flan-based models under the zero-shot setting (Table \ref{tab:zero-shot-flan-sst-2} in Appendix). We find that all models exhibit strong performance with minimal variance or even zero (e.g. Flan-T5-11B) between different positions. This outcome is expected due to the exposure of SST-2's training data during the instruction-tuning process, but it is also encouraging as it suggests that a strong language model should be robust to changes in prompt positioning. Nonetheless, we still observe a slight preference for the 'both' position in both Flan-T5-3B and Flan-LLaMA-13B models, supporting our hypothesis that one task might favour a prompt position more commonly encountered during the training process.


\section{General Discussion}\label{general_discussion}
Our main research question in this paper is whether and how prompt positions matter.  In Section \ref{gradient_based_experiments}, we observe that continuous prompts are more sensitive to prompt positions compared to manual prompts. In Section \ref{gradient_free_experiments}, all language models exhibit a certain degree of sensitivity to positional changing in both few-shot and zero-shot settings. Both sections \ref{gradient_based_experiments} and \ref{gradient_free_experiments} highlight that the optimal prompt position is not shared across tasks, and sometimes even differs among items of data. Furthermore, Section \ref{gradient_free_experiments} reveals that instruction-tuning fails to mitigate positional variance, possibly due to an unequal distribution of position variety in instruction templates. 

 Our goal is not to identify a 'best' position to replace the prompt position used in prior works. Rather, we aim to highlight the effect of prompt position, which is often overlooked before. For gradient-based prompting such as prompt tuning, due to the influence of training data items, we suggest instance-dependent prompt position optimisation as a valuable direction, with the potential to enhance model performance, which is similar to \citet{wu2022idpg,dynamicposition}.   As for gradient-free prompting, in real-world scenarios, prompt positioning and the logic in which questions or instructions are structured are diverse. A robust language model should display a consistent understanding of inputs with the same semantic essence when only the prompt position changes. A future work direction is to explore if increasing the position variety during the instruction tuning process improves the robustness of pre-trained models.  

\section{Conclusion} \label{sec: conclusion}
In this study, we evaluate the effect of prompt position in both zero-shot and few-shot scenarios. Our findings reveal significant variations in performance among different prompt positions across a variety of methods, tasks, and models. Additionally, our research indicates that prompt positions commonly adopted in the existing literature often result in sub-optimal performance, with no single prompt position universally excelling across all tasks. These findings suggest prompt position optimisation as a promising new direction in prompt engineering and advocate for the consideration of position-aware instruction tuning to develop more robust models in the future.

\section*{Limitations}
Due to the extensive workload of experiments, we only test our hypothesis for 5 sub-tasks from (\citet{gao2020LMBFF,lester2021power}) in gradient-based approaches and 10 sub-tasks from \citet{suzgun2022bbh} in gradient-free approaches. We use medium-sized language models in all our experiments (e.g. LLaMA-13B) using relatively low computational resources (i.e. a single GPU card for inferring or training), so although we strongly suspect the results will be similar in the latest Large Language Models (e.g. GPT4) this needs to be confirmed empirically in future experiments.

\section*{Acknowledgements}
This work was supported by the Economic and Social Research Council (ES/V011278/1). The authors acknowledge the use of the IRIDIS High Performance Computing Facility, and associated support services at the University of Southampton, in the completion of this work.

\bibliography{custom}

\begin{thebibliography}{43}
\expandafter\ifx\csname natexlab\endcsname\relax\def\natexlab#1{#1}\fi

\bibitem[{Ben{-}David et~al.(2021)Ben{-}David, Oved, and Reichart}]{PADA}
Eyal Ben{-}David, Nadav Oved, and Roi Reichart. 2021.
\newblock \href {http://arxiv.org/abs/2102.12206} {{PADA:} {A} prompt-based autoregressive approach for adaptation to unseen domains}.
\newblock \emph{CoRR}, abs/2102.12206.

\bibitem[{bench authors(2023)}]{srivastava2023beyond}
BIG bench authors. 2023.
\newblock \href {https://openreview.net/forum?id=uyTL5Bvosj} {Beyond the imitation game: Quantifying and extrapolating the capabilities of language models}.
\newblock \emph{Transactions on Machine Learning Research}.

\bibitem[{Brown et~al.(2020)Brown, Mann, Ryder, Subbiah, Kaplan, Dhariwal, Neelakantan, Shyam, Sastry, Askell, Agarwal, Herbert{-}Voss, Krueger, Henighan, Child, Ramesh, Ziegler, Wu, Winter, Hesse, Chen, Sigler, Litwin, Gray, Chess, Clark, Berner, McCandlish, Radford, Sutskever, and Amodei}]{GPT3}
Tom~B. Brown, Benjamin Mann, Nick Ryder, Melanie Subbiah, Jared Kaplan, Prafulla Dhariwal, Arvind Neelakantan, Pranav Shyam, Girish Sastry, Amanda Askell, Sandhini Agarwal, Ariel Herbert{-}Voss, Gretchen Krueger, Tom Henighan, Rewon Child, Aditya Ramesh, Daniel~M. Ziegler, Jeffrey Wu, Clemens Winter, Christopher Hesse, Mark Chen, Eric Sigler, Mateusz Litwin, Scott Gray, Benjamin Chess, Jack Clark, Christopher Berner, Sam McCandlish, Alec Radford, Ilya Sutskever, and Dario Amodei. 2020.
\newblock \href {http://arxiv.org/abs/2005.14165} {Language models are few-shot learners}.
\newblock \emph{CoRR}, abs/2005.14165.

\bibitem[{Chung et~al.(2022)Chung, Hou, Longpre, Zoph, Tay, Fedus, Li, Wang, Dehghani, Brahma, Webson, Gu, Dai, Suzgun, Chen, Chowdhery, Castro-Ros, Pellat, Robinson, Valter, Narang, Mishra, Yu, Zhao, Huang, Dai, Yu, Petrov, Chi, Dean, Devlin, Roberts, Zhou, Le, and Wei}]{chung2022scaling}
Hyung~Won Chung, Le~Hou, Shayne Longpre, Barret Zoph, Yi~Tay, William Fedus, Yunxuan Li, Xuezhi Wang, Mostafa Dehghani, Siddhartha Brahma, Albert Webson, Shixiang~Shane Gu, Zhuyun Dai, Mirac Suzgun, Xinyun Chen, Aakanksha Chowdhery, Alex Castro-Ros, Marie Pellat, Kevin Robinson, Dasha Valter, Sharan Narang, Gaurav Mishra, Adams Yu, Vincent Zhao, Yanping Huang, Andrew Dai, Hongkun Yu, Slav Petrov, Ed~H. Chi, Jeff Dean, Jacob Devlin, Adam Roberts, Denny Zhou, Quoc~V. Le, and Jason Wei. 2022.
\newblock \href {http://arxiv.org/abs/2210.11416} {Scaling instruction-finetuned language models}.

\bibitem[{Ding et~al.(2021)Ding, Hu, Zhao, Chen, Liu, Zheng, and Sun}]{ding2021openprompt}
Ning Ding, Shengding Hu, Weilin Zhao, Yulin Chen, Zhiyuan Liu, Hai-Tao Zheng, and Maosong Sun. 2021.
\newblock Openprompt: An open-source framework for prompt-learning.
\newblock \emph{arXiv preprint arXiv:2111.01998}.

\bibitem[{Gao et~al.(2021)Gao, Tow, Biderman, Black, DiPofi, Foster, Golding, Hsu, McDonell, Muennighoff, Phang, Reynolds, Tang, Thite, Wang, Wang, and Zou}]{evalharness}
Leo Gao, Jonathan Tow, Stella Biderman, Sid Black, Anthony DiPofi, Charles Foster, Laurence Golding, Jeffrey Hsu, Kyle McDonell, Niklas Muennighoff, Jason Phang, Laria Reynolds, Eric Tang, Anish Thite, Ben Wang, Kevin Wang, and Andy Zou. 2021.
\newblock \href {https://doi.org/10.5281/zenodo.5371628} {A framework for few-shot language model evaluation}.

\bibitem[{Gao et~al.(2020)Gao, Fisch, and Chen}]{gao2020LMBFF}
Tianyu Gao, Adam Fisch, and Danqi Chen. 2020.
\newblock Making pre-trained language models better few-shot learners.
\newblock \emph{arXiv preprint arXiv:2012.15723}.

\bibitem[{Gu et~al.(2022)Gu, Han, Liu, and Huang}]{gu2022ppt}
Yuxian Gu, Xu~Han, Zhiyuan Liu, and Minlie Huang. 2022.
\newblock \href {http://arxiv.org/abs/2109.04332} {Ppt: Pre-trained prompt tuning for few-shot learning}.

\bibitem[{Hu and Liu(2004)}]{CRhu2004mining}
Minqing Hu and Bing Liu. 2004.
\newblock Mining and summarizing customer reviews.
\newblock In \emph{Proceedings of the tenth ACM SIGKDD international conference on Knowledge discovery and data mining}, pages 168--177.

\bibitem[{Köksal et~al.(2023)Köksal, Schick, and Schütze}]{köksal2023meal}
Abdullatif Köksal, Timo Schick, and Hinrich Schütze. 2023.
\newblock \href {http://arxiv.org/abs/2211.08358} {Meal: Stable and active learning for few-shot prompting}.

\bibitem[{Lester et~al.(2021)Lester, Al-Rfou, and Constant}]{lester2021power}
Brian Lester, Rami Al-Rfou, and Noah Constant. 2021.
\newblock \href {http://arxiv.org/abs/2104.08691} {The power of scale for parameter-efficient prompt tuning}.

\bibitem[{Li and Liang(2021)}]{li-liang-2021-prefix}
Xiang~Lisa Li and Percy Liang. 2021.
\newblock \href {https://doi.org/10.18653/v1/2021.acl-long.353} {Prefix-tuning: Optimizing continuous prompts for generation}.
\newblock In \emph{Proceedings of the 59th Annual Meeting of the Association for Computational Linguistics and the 11th International Joint Conference on Natural Language Processing (Volume 1: Long Papers)}, pages 4582--4597, Online. Association for Computational Linguistics.

\bibitem[{Liu et~al.(2021{\natexlab{a}})Liu, Yuan, Fu, Jiang, Hayashi, and Neubig}]{promptsurveyLiu}
Pengfei Liu, Weizhe Yuan, Jinlan Fu, Zhengbao Jiang, Hiroaki Hayashi, and Graham Neubig. 2021{\natexlab{a}}.
\newblock \href {http://arxiv.org/abs/2107.13586} {Pre-train, prompt, and predict: {A} systematic survey of prompting methods in natural language processing}.
\newblock \emph{CoRR}, abs/2107.13586.

\bibitem[{Liu et~al.(2022)Liu, Sun, Huang, and Qiu}]{liu2022late}
Xiangyang Liu, Tianxiang Sun, Xuanjing Huang, and Xipeng Qiu. 2022.
\newblock \href {http://arxiv.org/abs/2210.11292} {Late prompt tuning: A late prompt could be better than many prompts}.

\bibitem[{Liu et~al.(2021{\natexlab{b}})Liu, Ji, Fu, Du, Yang, and Tang}]{ptuning2}
Xiao Liu, Kaixuan Ji, Yicheng Fu, Zhengxiao Du, Zhilin Yang, and Jie Tang. 2021{\natexlab{b}}.
\newblock \href {http://arxiv.org/abs/2110.07602} {P-tuning v2: Prompt tuning can be comparable to fine-tuning universally across scales and tasks}.
\newblock \emph{CoRR}, abs/2110.07602.

\bibitem[{Liu et~al.(2021{\natexlab{c}})Liu, Zheng, Du, Ding, Qian, Yang, and Tang}]{P-tuning}
Xiao Liu, Yanan Zheng, Zhengxiao Du, Ming Ding, Yujie Qian, Zhilin Yang, and Jie Tang. 2021{\natexlab{c}}.
\newblock \href {http://arxiv.org/abs/2103.10385} {{GPT} understands, too}.
\newblock \emph{CoRR}, abs/2103.10385.

\bibitem[{Liu et~al.(2019)Liu, Ott, Goyal, Du, Joshi, Chen, Levy, Lewis, Zettlemoyer, and Stoyanov}]{roberta}
Yinhan Liu, Myle Ott, Naman Goyal, Jingfei Du, Mandar Joshi, Danqi Chen, Omer Levy, Mike Lewis, Luke Zettlemoyer, and Veselin Stoyanov. 2019.
\newblock \href {http://arxiv.org/abs/1907.11692} {Roberta: {A} robustly optimized {BERT} pretraining approach}.
\newblock \emph{CoRR}, abs/1907.11692.

\bibitem[{Loshchilov and Hutter(2017)}]{adamw}
Ilya Loshchilov and Frank Hutter. 2017.
\newblock \href {http://arxiv.org/abs/1711.05101} {Fixing weight decay regularization in adam}.
\newblock \emph{CoRR}, abs/1711.05101.

\bibitem[{Lu et~al.(2021)Lu, Bartolo, Moore, Riedel, and Stenetorp}]{lu2021fantastically}
Yao Lu, Max Bartolo, Alastair Moore, Sebastian Riedel, and Pontus Stenetorp. 2021.
\newblock Fantastically ordered prompts and where to find them: Overcoming few-shot prompt order sensitivity.
\newblock \emph{arXiv preprint arXiv:2104.08786}.

\bibitem[{Mao et~al.(2022)Mao, Liu, He, Li, and Cambria}]{Maoposition}
Rui Mao, Qian Liu, Kai He, Wei Li, and Erik Cambria. 2022.
\newblock \href {https://doi.org/10.1109/TAFFC.2022.3204972} {The biases of pre-trained language models: An empirical study on prompt-based sentiment analysis and emotion detection}.
\newblock \emph{IEEE Transactions on Affective Computing}, pages 1--11.

\bibitem[{Prasad et~al.(2022)Prasad, Hase, Zhou, and Bansal}]{prasad2022grips}
Archiki Prasad, Peter Hase, Xiang Zhou, and Mohit Bansal. 2022.
\newblock Grips: Gradient-free, edit-based instruction search for prompting large language models.
\newblock \emph{arXiv preprint arXiv:2203.07281}.

\bibitem[{Raffel et~al.(2023)Raffel, Shazeer, Roberts, Lee, Narang, Matena, Zhou, Li, and Liu}]{raffel2023exploring}
Colin Raffel, Noam Shazeer, Adam Roberts, Katherine Lee, Sharan Narang, Michael Matena, Yanqi Zhou, Wei Li, and Peter~J. Liu. 2023.
\newblock \href {http://arxiv.org/abs/1910.10683} {Exploring the limits of transfer learning with a unified text-to-text transformer}.

\bibitem[{Sanh et~al.(2022)Sanh, Webson, Raffel, Bach, Sutawika, Alyafeai, Chaffin, Stiegler, Scao, Raja, Dey, Bari, Xu, Thakker, Sharma, Szczechla, Kim, Chhablani, Nayak, Datta, Chang, Jiang, Wang, Manica, Shen, Yong, Pandey, Bawden, Wang, Neeraj, Rozen, Sharma, Santilli, Fevry, Fries, Teehan, Bers, Biderman, Gao, Wolf, and Rush}]{sanh2022multitask}
Victor Sanh, Albert Webson, Colin Raffel, Stephen~H. Bach, Lintang Sutawika, Zaid Alyafeai, Antoine Chaffin, Arnaud Stiegler, Teven~Le Scao, Arun Raja, Manan Dey, M~Saiful Bari, Canwen Xu, Urmish Thakker, Shanya~Sharma Sharma, Eliza Szczechla, Taewoon Kim, Gunjan Chhablani, Nihal Nayak, Debajyoti Datta, Jonathan Chang, Mike Tian-Jian Jiang, Han Wang, Matteo Manica, Sheng Shen, Zheng~Xin Yong, Harshit Pandey, Rachel Bawden, Thomas Wang, Trishala Neeraj, Jos Rozen, Abheesht Sharma, Andrea Santilli, Thibault Fevry, Jason~Alan Fries, Ryan Teehan, Tali Bers, Stella Biderman, Leo Gao, Thomas Wolf, and Alexander~M. Rush. 2022.
\newblock \href {http://arxiv.org/abs/2110.08207} {Multitask prompted training enables zero-shot task generalization}.

\bibitem[{Schick and Sch{\"u}tze(2020)}]{schick2020IPET}
Timo Schick and Hinrich Sch{\"u}tze. 2020.
\newblock Exploiting cloze questions for few shot text classification and natural language inference.
\newblock \emph{arXiv preprint arXiv:2001.07676}.

\bibitem[{Schick and Sch{\"{u}}tze(2020)}]{IPET2}
Timo Schick and Hinrich Sch{\"{u}}tze. 2020.
\newblock \href {http://arxiv.org/abs/2009.07118} {It's not just size that matters: Small language models are also few-shot learners}.
\newblock \emph{CoRR}, abs/2009.07118.

\bibitem[{See et~al.(2017)See, Liu, and Manning}]{CNN-DM-see-etal-2017-get}
Abigail See, Peter~J. Liu, and Christopher~D. Manning. 2017.
\newblock \href {https://doi.org/10.18653/v1/P17-1099} {Get to the point: Summarization with pointer-generator networks}.
\newblock In \emph{Proceedings of the 55th Annual Meeting of the Association for Computational Linguistics (Volume 1: Long Papers)}, pages 1073--1083, Vancouver, Canada. Association for Computational Linguistics.

\bibitem[{Shazeer and Stern(2018)}]{shazeer2018adafactor}
Noam Shazeer and Mitchell Stern. 2018.
\newblock Adafactor: Adaptive learning rates with sublinear memory cost.
\newblock In \emph{International Conference on Machine Learning}, pages 4596--4604. PMLR.

\bibitem[{Shin et~al.(2020)Shin, Razeghi, Logan~IV, Wallace, and Singh}]{shin2020autoprompt}
Taylor Shin, Yasaman Razeghi, Robert~L Logan~IV, Eric Wallace, and Sameer Singh. 2020.
\newblock Autoprompt: Eliciting knowledge from language models with automatically generated prompts.
\newblock \emph{arXiv preprint arXiv:2010.15980}.

\bibitem[{Sun et~al.(2022)Sun, Shao, Qian, Huang, and Qiu}]{sun2022blackbox}
Tianxiang Sun, Yunfan Shao, Hong Qian, Xuanjing Huang, and Xipeng Qiu. 2022.
\newblock \href {http://arxiv.org/abs/2201.03514} {Black-box tuning for language-model-as-a-service}.

\bibitem[{Suzgun et~al.(2022)Suzgun, Scales, Schärli, Gehrmann, Tay, Chung, Chowdhery, Le, Chi, Zhou, and Wei}]{suzgun2022bbh}
Mirac Suzgun, Nathan Scales, Nathanael Schärli, Sebastian Gehrmann, Yi~Tay, Hyung~Won Chung, Aakanksha Chowdhery, Quoc~V. Le, Ed~H. Chi, Denny Zhou, and Jason Wei. 2022.
\newblock \href {http://arxiv.org/abs/2210.09261} {Challenging big-bench tasks and whether chain-of-thought can solve them}.

\bibitem[{Touvron et~al.(2023)Touvron, Lavril, Izacard, Martinet, Lachaux, Lacroix, Rozière, Goyal, Hambro, Azhar, Rodriguez, Joulin, Grave, and Lample}]{touvron2023llama}
Hugo Touvron, Thibaut Lavril, Gautier Izacard, Xavier Martinet, Marie-Anne Lachaux, Timothée Lacroix, Baptiste Rozière, Naman Goyal, Eric Hambro, Faisal Azhar, Aurelien Rodriguez, Armand Joulin, Edouard Grave, and Guillaume Lample. 2023.
\newblock \href {http://arxiv.org/abs/2302.13971} {Llama: Open and efficient foundation language models}.

\bibitem[{Voorhees and Tice(2000)}]{TREC}
Ellen~M Voorhees and Dawn~M Tice. 2000.
\newblock Building a question answering test collection.
\newblock In \emph{Proceedings of the 23rd annual international ACM SIGIR conference on Research and development in information retrieval}, pages 200--207.

\bibitem[{Wang et~al.(2019)Wang, Pruksachatkun, Nangia, Singh, Michael, Hill, Levy, and Bowman}]{wang2019superglue}
Alex Wang, Yada Pruksachatkun, Nikita Nangia, Amanpreet Singh, Julian Michael, Felix Hill, Omer Levy, and Samuel Bowman. 2019.
\newblock Superglue: A stickier benchmark for general-purpose language understanding systems.
\newblock \emph{Advances in neural information processing systems}, 32.

\bibitem[{Wang et~al.(2018)Wang, Singh, Michael, Hill, Levy, and Bowman}]{wang2018glue}
Alex Wang, Amanpreet Singh, Julian Michael, Felix Hill, Omer Levy, and Samuel~R Bowman. 2018.
\newblock Glue: A multi-task benchmark and analysis platform for natural language understanding.
\newblock \emph{arXiv preprint arXiv:1804.07461}.

\bibitem[{Wang et~al.(2023)Wang, Ivison, Dasigi, Hessel, Khot, Chandu, Wadden, MacMillan, Smith, Beltagy, and Hajishirzi}]{flanllama}
Yizhong Wang, Hamish Ivison, Pradeep Dasigi, Jack Hessel, Tushar Khot, Khyathi~Raghavi Chandu, David Wadden, Kelsey MacMillan, Noah~A. Smith, Iz~Beltagy, and Hannaneh Hajishirzi. 2023.
\newblock \href {http://arxiv.org/abs/2306.04751} {How far can camels go? exploring the state of instruction tuning on open resources}.

\bibitem[{Wei et~al.(2022{\natexlab{a}})Wei, Bosma, Zhao, Guu, Yu, Lester, Du, Dai, and Le}]{wei2022instruction}
Jason Wei, Maarten Bosma, Vincent~Y. Zhao, Kelvin Guu, Adams~Wei Yu, Brian Lester, Nan Du, Andrew~M. Dai, and Quoc~V. Le. 2022{\natexlab{a}}.
\newblock \href {http://arxiv.org/abs/2109.01652} {Finetuned language models are zero-shot learners}.

\bibitem[{Wei et~al.(2022{\natexlab{b}})Wei, Wang, Schuurmans, Bosma, Xia, Chi, Le, Zhou et~al.}]{wei2022chain}
Jason Wei, Xuezhi Wang, Dale Schuurmans, Maarten Bosma, Fei Xia, Ed~Chi, Quoc~V Le, Denny Zhou, et~al. 2022{\natexlab{b}}.
\newblock Chain-of-thought prompting elicits reasoning in large language models.
\newblock \emph{Advances in Neural Information Processing Systems}, 35:24824--24837.

\bibitem[{Wu et~al.(2022)Wu, Wang, Gu, Hou, Dong, Vydiswaran, and Ma}]{wu2022idpg}
Zhuofeng Wu, Sinong Wang, Jiatao Gu, Rui Hou, Yuxiao Dong, VG~Vydiswaran, and Hao Ma. 2022.
\newblock Idpg: An instance-dependent prompt generation method.
\newblock \emph{arXiv preprint arXiv:2204.04497}.

\bibitem[{Yang et~al.(2023)Yang, Cheng, Zhao, Petzold, and Chen}]{dynamicposition}
Xianjun Yang, Wei Cheng, Xujiang Zhao, Linda Petzold, and Haifeng Chen. 2023.
\newblock \href {http://arxiv.org/abs/2303.02909} {Dynamic prompting: A unified framework for prompt tuning}.

\bibitem[{Yao et~al.(2023)Yao, Yu, Zhao, Shafran, Griffiths, Cao, and Narasimhan}]{yao2023tree}
Shunyu Yao, Dian Yu, Jeffrey Zhao, Izhak Shafran, Thomas~L Griffiths, Yuan Cao, and Karthik Narasimhan. 2023.
\newblock Tree of thoughts: Deliberate problem solving with large language models.
\newblock \emph{arXiv preprint arXiv:2305.10601}.

\bibitem[{Zhang et~al.(2022)Zhang, Zhang, Li, and Smola}]{zhang2022autocot}
Zhuosheng Zhang, Aston Zhang, Mu~Li, and Alex Smola. 2022.
\newblock \href {http://arxiv.org/abs/2210.03493} {Automatic chain of thought prompting in large language models}.

\bibitem[{Zhao et~al.(2021)Zhao, Wallace, Feng, Klein, and Singh}]{zhao2021calibrate}
Zihao Zhao, Eric Wallace, Shi Feng, Dan Klein, and Sameer Singh. 2021.
\newblock Calibrate before use: Improving few-shot performance of language models.
\newblock In \emph{International Conference on Machine Learning}, pages 12697--12706. PMLR.

\bibitem[{Zhou et~al.(2022)Zhou, Muresanu, Han, Paster, Pitis, Chan, and Ba}]{zhou2022largemodelsarehuman}
Yongchao Zhou, Andrei~Ioan Muresanu, Ziwen Han, Keiran Paster, Silviu Pitis, Harris Chan, and Jimmy Ba. 2022.
\newblock Large language models are human-level prompt engineers.
\newblock \emph{arXiv preprint arXiv:2211.01910}.

\end{thebibliography}

\appendix
\clearpage
\section{The Effect of [Mask] Token Position}\label{appendix:mask_token}
We have discovered that the position of the [mask] token has an impact on the cloze-style prompt, namely within Masked Language Models. To investigate this further, we conduct null template experiments with a K size of 16, where we simply concatenate the inputs and the [mask] token without a prompt. By analyzing the results of the null template as presented in Table \ref{tab:null_template}, we observe that in single-sentence tasks, placing text before the [mask] token generally leads to better performance. For sentence-pair tasks, placing [mask] before the text is relatively sub-optimal. Notably, for the RTE task, positioning [mask] token in the middle of the two original inputs proves to be more advantageous. This observation also aligns with the overall performance trend demonstrated in the complete set of results, which can be found in Appendix \ref{appendix: all results}. 

Interestingly, even when the [mask] token is placed in the same relative position to the task inputs, the performance exhibits a noticeable difference depending on how to insert the prompt sequences. For instance, in the cloze continuous prompt experiment on RTE dataset with a K size of 128, when the [mask] token is placed in the middle of two task inputs, "\textit{\{text\_a\} [mask] \{text\_b\}  \textit{\textbf{P}}}" achieves a performance that is 6.28 higher compared to "\textit{\{text\_a\} \textit{\textbf{P}} [mask] \{text\_b\}}" (detailed in Table \ref{tab:CC-RTE}). 


\begin{table}[ht!]
\resizebox{0.45\textwidth}{!}{
\begin{tabular}{lll}
    \hline\noalign{\smallskip}
Dataset       & Null template              & Mean (std)     \\
    \hline\noalign{\smallskip}
SST-2         & \{text\_a\}  {[}mask{]}       & 89.40 (1.44)  \\
              & {[}mask{]}   \{text\_a\}      & 82.80 (5.56)  \\
        \hline\noalign{\smallskip}
CR            & \{text\_a\}  {[}mask{]}       & 90.66 (1.19)  \\
              & {[}mask{]}   \{text\_a\}      & 89.01 (1.76) \\
            \hline\noalign{\smallskip}
TREC & \{text\_a\}  {[}mask{]}       & 85.99 (1.82)  \\
              & {[}mask{]}   \{text\_a\}      & 83.58 (1.58)  \\
        \hline\noalign{\smallskip}
RTE           & \{text\_a\} \{text\_b\} {[}mask{]} & 55.02 (7.16)  \\
              & \{text\_a\} {[}mask{]} \{text\_b\} & 65.49 (3.58)  \\
              & {[}mask{]} \{text\_a\} \{text\_b\} & 53.86 (5.58) \\
        \hline\noalign{\smallskip}
BoolQ         & \{text\_a\} \{text\_b\} {[}mask{]} & 64.70 (3.09) \\
              & \{text\_a\} {[}mask{]} \{text\_b\} & 64.05 (4.13)  \\
              & {[}mask{]} \{text\_a\} \{text\_b\} & 59.05 (1.57)  \\
        \hline\noalign{\smallskip}
\end{tabular}
}
  \caption{Null template results for all datasets on K size of 16 for cloze manual prompt.}
  \label{tab:null_template}
\end{table}

\section{Experimental Details}
For prompt-based fine-tuning, we employ an AdamW optimizer \cite{adamw} with a learning rate of 2e-5 and a batch size of 8 for 1000 steps, validating the performance every 100 steps. For prompt tuning on the Roberta model, we follow the setting in \citet{sun2022blackbox}, using AdamW with a learning rate of 5e-4 and a batch size of 16 for 1000 epochs, with model performance validation every 100 steps. For prompt tuning on the T5 model, we adopt Adafactor \cite{shazeer2018adafactor} with a learning rate of 0.3 and a batch size of 16 for 1000 steps, evaluating the performance every 8 steps. The prompt length for all experiments is set to 50, initialized from the first 50 tokens embeddings of the pre-trained language model following \citet{ding2021openprompt} as initializing from the language model's vocabulary often gives better results. To mitigate overfitting, we employ the strategy of early stopping across all experiments. 

All our models were trained on a single RTX8000 with 48GB of memory. In the main training experiments, continuous prompt tuning took approximately 547 hours for T5-3B, 264 hours for T5-Large, and 2269 hours for RoBERTa-large; prompt-based fine-tuning took around 150 hours for T5-Large and 546 hours for RoBERTa-large. For the gradient-free main experiments, zero-shot experiments took roughly 16 hours, few-shot direct prompting took about 70 hours, and CoT prompting took around 345 hours.


\section{Datasets, Reference Prompts and Positions}
\label{appendix: reference prompt}
For gradient-based approaches, the dataset statistics used are shown in Table \ref{tab:dataset}, the reference prompt positions are shown in Table \ref{tab:reference_template}, and the best-performing prompt positions for single-tasks are shown in Table \ref{tab:best-single}. The CR templates employed here are consistent with that of SST-2 for all methods, following the setting outlined in \citet{gao2020LMBFF}. Regarding the prefix continuous prompt (PC) applied to the TREC dataset, we follow the prompt position setting provided by \citet{lester2021power}, which is commonly used as the default prompt position for most continuous prompt methods. Besides, \citet{gu2022ppt} use "\textit{\textbf{\textit{P}} \{text\_b\} [mask] \{text\_a\}}" (e.g. \{text\_a\} is "passage" and \{text\_b\} is "question" in BoolQ task) as the prompt position in cloze continuous prompt method. To narrow our focus on the prompt position and ensure consistency with the expected task input sequence order in other methods (e.g. manual prompt), we modify their input orders in this specific case. For gradient-free approaches, reference prompt positions used in \citet{suzgun2022bbh} and their variants are shown in Table \ref{tab:reference_position_BBH}.

\begin{table*}[htbp]
  \centering
    \resizebox{0.7\textwidth}{!}{
    \begin{tabular}{lrrllr}
     \hline\noalign{\smallskip}   
    Corpus & $|Y|$ & \multicolumn{1}{l}{Train} & \multicolumn{1}{l}{Validation} & Task  & Evaluation Metrics \\
    \hline\noalign{\smallskip}
    Single Sentence Tasks &       &       &       &       &  \\
    \hline\noalign{\smallskip}
    CR    & 2 & 1775  & 2000  & sentiment & accuracy \\
    SST-2 & 2 & 67349 & 1821  & sentiment & accuracy  \\
      TREC & 6 & 5452 & 500 & question cls.& accuracy  \\
    \hline\noalign{\smallskip}
    Sentence Pair Tasks &       &       &       &       &  \\
    \hline\noalign{\smallskip}
    RTE   & 2 & 2491  & 278   & NLI   & accuracy  \\
    BoolQ & 2 & 9427  & 3270  & QA    & accuracy  \\
    \hline\noalign{\smallskip}
    \end{tabular}%
    }
 \caption{The datasets evaluated in our work. $|Y|$ represents the number of classes within each task. We only sample $D_{train}$ and $D_{dev}$ of $K\times|Y|$ examples from the original training set in the few-shot experiments.}
  \label{tab:dataset}%
\end{table*}%

\begin{table*}[htbp!]
\centering
\resizebox{\textwidth}{!}{
\begin{tabular}{llp{6cm}p{6cm}l}
\hline\noalign{\smallskip}
\textbf{Method} & \textbf{Task} & \textbf{Template} & \textbf{Verbalizer} & \textbf{Reference} \\ 
\hline\noalign{\smallskip}
\textbf{CM} & SST-2 & \{text\_a\} It was [mask]. & positive: great, negative: terrible & \cite{gao2020LMBFF} \\ 
 & CR & \{text\_a\} It was [mask]. & positive: great, negative: terrible & \cite{gao2020LMBFF} \\
& TREC & \{text\_a\} This question is related to [mask] category. & abbr.: Expression, entity: Entity, description: Description, human: Human, location: Location, numeric: Number & \cite{köksal2023meal} \\
 & RTE & \{text\_a\} Question: \{text\_b\}? the Answer: [mask]. & entailment: yes, not\_entailment: no & \cite{P-tuning} \\
 & BoolQ & \{text\_a\}. Question: \{text\_b\}? Answer: [mask]. & entailment: yes, not\_entailment: no & \cite{IPET2} \\
\hline\noalign{\smallskip}
\textbf{CC} & SST-2 & \textbf{\textit{P}} \{text\_a\}  [mask] & positive: great, negative: terrible & \cite{gu2022ppt} \\
 & CR & \textbf{\textit{P}}  \{text\_a\}  [mask] & positive: great, negative: terrible &  \\
 & TREC & \textbf{\textit{P}}  [mask]   \{text\_a\} & abbr.: Expression, entity: Entity, description: Description, human: Human, location: Location, numeric: Number & \cite{liu2022late} \\
 
 & RTE & \textbf{\textit{P}}  \{text\_a\}  [mask]  \{text\_b\} & entailment: yes, not\_entailment: no & \cite{gu2022ppt} \\
 & BoolQ & \textbf{\textit{P}}  \{text\_a\}  [mask]  \{text\_b\} & true: yes, false: no & \cite{gu2022ppt}\\
\hline\noalign{\smallskip}
\textbf{PM} & SST-2 & \{text\_a\} Question: Is this sentence positive or negative? Answer: & positive: positive, negative: negative & \cite{evalharness} \\
 & CR & \{text\_a\} Question: Is this sentence positive or negative? Answer: & positive: positive, negative: negative &  \\
 &TREC & Categories: $\{', '.join(label\_words)\}$. What category best describes: \{text\_a\} Answer: & abbr.: Abbreviation, entity: Entity, description: Description, human: Person, location: Location, numeric: Quantity &\cite{sanh2022multitask} \\
 & RTE & \{text\_a\} Question: \{text\_b\} True or False? Answer: & entailment: true, not\_entailment: false & \cite{GPT3} \\
 & BoolQ & \{text\_a\} Question: \{text\_b\}? Answer: & true: yes, false: no & \cite{GPT3} \\
\hline\noalign{\smallskip}
\textbf{PC} & SST-2 & \textbf{\textit{P}}  \{text\_a\} & positive: positive, negative: negative & \cite{lester2021power} \\
 & CR & \textbf{\textit{P}}  \{text\_a\} & positive: positive, negative: negative & \\
 & TREC & \textbf{\textit{P}}  \{text\_a\} & abbr.: Abbreviation, entity: Entity, description: Description, human: Person, location: Location, numeric: Quantity & \\
 & RTE & \textbf{\textit{P}}  \{text\_a\}  \{text\_b\} & entailment: entailment, not\_entailment: contradiction & \cite{lester2021power} \\
 & BoolQ & \textbf{\textit{P}}  \{text\_a\}  \{text\_b\} & true: true, false: false & \cite{lester2021power} \\
\hline\noalign{\smallskip}
\end{tabular}
}
\caption{All reference prompt positions used in the main text of the paper. \textit{\textbf{P}} denotes continuous prompt tokens.}
\label{tab:reference_template}
\end{table*}

\begin{table*}[ht]
\centering
\resizebox{\textwidth}{!}{
\begin{tabular}{llp{6cm}p{6cm}p{8cm}}
\hline\noalign{\smallskip}
K-Shot & Method & SST-2 & CR  & TREC  \\
\hline\noalign{\smallskip}
K=16   & CM     & \{text\_a\} It was {[}mask{]}.                                & {[}mask{]} It was \{text\_a\}.                                 & \{text\_a\} {[}mask{]} This question is related to category.                                              \\
& CC     & \textit{\textbf{P}} {[}mask{]} \textit{\textbf{P}} \{text\_a\}                                 
       & \textit{\textbf{P}} \{text\_a\} \textit{\textbf{P}} {[}mask{]}    & \textit{\textbf{P}} \{text\_a\} {[}mask{]}                                                       \\
       & PM     & \{text\_a\} \textit{Q}: Is this sentence positive or negative?  \textit{A}:                         
                & \textit{Q}: Is this sentence positive or negative?  \{text\_a\} \textit{A}:                          
            & \textit{A}: \textit{C}: $\{', '.join(label\_words)\}$. 
            What category best describes:   \{text\_a\}  \\
\textbf{} & PC    & \{text\_a\} \textit{\textbf{P}}                            & \textit{\textbf{P}} \{text\_a\}                             & \textit{\textbf{P}} \{text\_a\} \textit{\textbf{P}}   
\\
\textbf{} & PC-XL    & \textit{\textbf{P}} \{text\_a\}                             & \textit{\textbf{P}} \{text\_a\}                              & \{text\_a\} \textit{\textbf{P}}   \\  

\hline\noalign{\smallskip}
K=32   & CM     & It was {[}mask{]}. \{text\_a\}                   & \{text\_a\}. It was {[}mask{]}                   & This question is related to category. \{text\_a\} {[}mask{]}                                                                      \\
       & CC     & \textit{\textbf{P}} \{text\_a\} {[}mask{]}    
 & {[}mask{]} \textit{\textbf{P}} \{text\_a\}                  & \textit{\textbf{P}} {[}mask{]} \{text\_a\} \textit{\textbf{P}}            \\
       & PM     & \{text\_a\} \textit{Q}: Is this sentence positive or negative?  \textit{A}:                           
                & \textit{Q}: Is this sentence positive or negative?  \{text\_a\} \textit{A}:                         
                & \{text\_a\} \textit{C}: $\{', '.join(label\_words)\}$. What category best describes: \textit{A}:    \\
       & PC     & \{text\_a\} \textit{\textbf{P}}                           & \textit{\textbf{P}} \{text\_a\}                  & \textit{\textbf{P}} \{text\_a\} \textit{\textbf{P}}                                           \\
       \textbf{} & PC-XL    & \textit{\textbf{P}} \{text\_a\}                             & \textit{\textbf{P}} \{text\_a\} \textit{\textbf{P}}                             & \{text\_a\} \textit{\textbf{P}}   \\ 
\hline\noalign{\smallskip}
K=64   & CM     & {[}mask{]} It was \{text\_a\}.                    & It was. \{text\_a\} {[}mask{]}                    & \{text\_a\} {[}mask{]} This question is related to category.                                                \\
       & CC     & \textit{\textbf{P}} {[}mask{]}  \textit{\textbf{P}} \{text\_a\}                                 
    & \{text\_a\} {[}mask{]} \textit{\textbf{P}}                           
    & \textit{\textbf{P}} {[}mask{]} \{text\_a\}     \\
\textbf{} & PM     & \textit{Q}: Is this sentence positive or negative? \textit{A}: \{text\_a\}                      & \{text\_a\}\textit{Q}: Is this sentence positive or negative?  \textit{A}:                    & \textit{C}: $\{', '.join(label\_words)\}$. What category best describes: \{text\_a\} \textit{A}:    \\
       & PC     & \{text\_a\} \textit{\textbf{P}}       & \textit{\textbf{P}} \{text\_a\}    & \{text\_a\} \textit{\textbf{P}}    \\
       
       \textbf{} & PC-XL    & \textit{\textbf{P}} \{text\_a\}                             & \textit{\textbf{P}} \{text\_a\} \textit{\textbf{P}}                             & \textit{\textbf{P}} \{text\_a\}   \\ 
\hline\noalign{\smallskip}
K=128  & CM     & It was {[}mask{]}. \{text\_a\}                          
                & \{text\_a\} It was {[}mask{]}.   & \{text\_a\} {[}mask{]} This question is related to category.   \\
& CC     & \textit{\textbf{P}} {[}mask{]} \textit{\textbf{P}} \{text\_a\}          & \textit{\textbf{P}} \{text\_a\} \textit{\textbf{P}} {[}mask{]}           & \textit{\textbf{P}} \{text\_a\} {[}mask{]} \textit{\textbf{P}}        \\
       & PM     & \textit{Q}: Is this sentence positive or negative? \textit{A}: \{text\_a\}                           
                & \textit{Q}: Is this sentence positive or negative? \textit{A}: \{text\_a\}                           
                & \textit{A}: \textit{C}: $\{', '.join(label\_words)\}$. What category best describes: \{text\_a\}     \\
    & PC     & \textit{\textbf{P}} \{text\_a\}             & \textit{\textbf{P}} \{text\_a\}  
    & \textit{\textbf{P}} \{text\_a\} \textit{\textbf{P}}     \\
    
    \textbf{} & PC-XL    & \textit{\textbf{P}} \{text\_a\}                             & \textit{\textbf{P}} \{text\_a\}                              & \textit{\textbf{P}} \{text\_a\} \textit{\textbf{P}}   \\ 
\hline\noalign{\smallskip}
\end{tabular}
}
  \caption{The best-performing prompt positions for single-sentence tasks on SST-2, CR and TREC datasets. \textit{\textbf{P}} denotes continuous prompt tokens, while \textit{Q}, \textit{A}, and \textit{C} represent the abbreviations for "Question", "Answer", and "Categories", respectively. $\{', '.join(label\_words)\}$ simplifies the representation of six label words used in the TREC dataset.}
  \label{tab:best-single}%
\end{table*}

\begin{table*}[ht!]
\centering
\resizebox{0.9\textwidth}{!}{
\begin{tabular}{llp{0.9\textwidth}}
\hline\noalign{\smallskip}
Causal Judgement &
  F &
  Q: How would a typical person answer each of the following questions about causation?\textbackslash{}n\{Options\}\textbackslash{}n\{Input\}\textbackslash{}nA: 
   \\
 &
  B* &
  Q: How would a typical person answer each of the following questions about causation?\textbackslash{}n\{Input\}\textbackslash{}n\{Options\}\textbackslash{}nA:
   \\
 &
  R &
  Q: \{Input\}\textbackslash{}n\{Options\}\textbackslash{}nHow would a typical person answer each of the following questions about causation?\textbackslash{}nA: 
   \\
   \hline\noalign{\smallskip}
Disambiguation QA &
  F &
  Q: In the following sentences, explain the antecedent of the pronoun (which thing the pronoun refers to), or state that it is ambiguous.\textbackslash{}n\{Options\}\textbackslash{}n\{Input\}\textbackslash{}nA: 
   \\
 &
  B* &
  Q: In the following sentences, explain the antecedent of the pronoun (which thing the pronoun refers to), or state that it is ambiguous.\textbackslash{}n\{Input\}\textbackslash{}n\{Options\}\textbackslash{}nA: 
   \\
 &
  R &
  Q: \{Input\}\textbackslash{}nIn the following sentences, explain the antecedent of the pronoun (which thing the pronoun refers to), or state that it is ambiguous.\textbackslash{}n\{Options\}\textbackslash{}nA: 
   \\
   \hline\noalign{\smallskip}
Sports Understanding &
  F* &
  Q: Is the following sentence plausible? \{Input\}\textbackslash{}nA: 
   \\
 &
  B &
  Q: Is the following sentence \{Input\} plausible?\textbackslash{}nA: 
   \\
 &
  R &
  Q: \{Input\} Is the following sentence plausible?\textbackslash{}nA: 
   \\
   \hline\noalign{\smallskip}
Navigate &
  F &
  Q: If you follow these instructions, do you return to the starting point?\textbackslash{}n\{Options\}\textbackslash{}n\{Input\}\textbackslash{}nA: 
   \\
 &
  B* &
  Q: If you follow these instructions, do you return to the starting point? \{Input\}\textbackslash{}n\{Options\}\textbackslash{}nA: 
   \\
 &
  R &
  Q: \{Input\} If you follow these instructions, do you return to the starting point?\textbackslash{}n\{Options\}\textbackslash{}nA: 
   \\
   \hline\noalign{\smallskip}
Logical Deduction (3,5,7) &
  F &
  Q: The following paragraphs each describe a set of three (five/seven) objects arranged in a fixed order. The statements are logically consistent within each paragraph.\textbackslash{}n\{Options\}\textbackslash{}n\{Input\}\textbackslash{}nA: 
   \\
 &
  B* &
  Q: The following paragraphs each describe a set of three (five/seven) objects arranged in a fixed order. The statements are logically consistent within each paragraph. \{Input\}\textbackslash{}n\{Options\}\textbackslash{}nA: 
   \\
 &
  R &
  Q: \{Input\} The following paragraphs each describe a set of three (five/seven) objects arranged in a fixed order. The statements are logically consistent within each paragraph.\textbackslash{}n\{Options\}\textbackslash{}nA: 
   \\
   \hline\noalign{\smallskip}
Penguins in a Table &
  F &
  Q: Here is a table where the first line is a header and each subsequent line is a penguin:  name, age, height (cm), weight (kg) Louis, 7, 50, 11 Bernard, 5, 80, 13 Vincent, 9, 60, 11 Gwen, 8, 70, 15  For example: the age of Louis is 7, the weight of Gwen is 15 kg, the height of Bernard is 80 cm.\textbackslash{}n\{Options\}\textbackslash{}n\{Input\}\textbackslash{}nA: 
   \\
 &
  B* &
  Q: Here is a table where the first line is a header and each subsequent line is a penguin:  name, age, height (cm), weight (kg) Louis, 7, 50, 11 Bernard, 5, 80, 13 Vincent, 9, 60, 11 Gwen, 8, 70, 15  For example: the age of Louis is 7, the weight of Gwen is 15 kg, the height of Bernard is 80 cm.  \{Input\}\textbackslash{}n\{Options\}\textbackslash{}nA: 
   \\
 &
  R &
  Q: \{Input\}  Here is a table where the first line is a header and each subsequent line is a penguin:  name, age, height (cm), weight (kg) Louis, 7, 50, 11 Bernard, 5, 80, 13 Vincent, 9, 60, 11 Gwen, 8, 70, 15  For example: the age of Louis is 7, the weight of Gwen is 15 kg, the height of Bernard is 80 cm.\textbackslash{}n\{Options\}\textbackslash{}nA: 
   \\
   \hline\noalign{\smallskip}
Salient Translation Err. &
  F &
  Q: The following translations from German to English contain a particular error. That error will be one of the following types: Named Entities: An entity (names, places, locations, etc.) is changed to a different entity. Numerical Values: Numerical values (ordinals or cardinals), dates, and/or units are changed. Modifiers or Adjectives: The modifiers and adjectives pertaining to a noun are changed. Negation or Antonyms: Introduce or remove a negation or change comparatives to their antonyms. Facts: Trivial factual errors not pertaining to the above classes are introduced in the translations. Dropped Content: A significant clause in the translation is removed. Please identify that error.\textbackslash{}n\{Options\}\textbackslash{}n\{Input\}\textbackslash{}nA: 
   \\
 &
  B* &
  Q: The following translations from German to English contain a particular error. That error will be one of the following types: Named Entities: An entity (names, places, locations, etc.) is changed to a different entity. Numerical Values: Numerical values (ordinals or cardinals), dates, and/or units are changed. Modifiers or Adjectives: The modifiers and adjectives pertaining to a noun are changed. Negation or Antonyms: Introduce or remove a negation or change comparatives to their antonyms. Facts: Trivial factual errors not pertaining to the above classes are introduced in the translations. Dropped Content: A significant clause in the translation is removed. Please identify that error.  \{Input\}\textbackslash{}n\{Options\}\textbackslash{}nA: 
   \\
 &
  R &
  Q: \{Input\}\textbackslash{}n\{Options\}\textbackslash{}nThe following translations from German to English contain a particular error. That error will be one of the following types: Named Entities: An entity (names, places, locations, etc.) is changed to a different entity. Numerical Values: Numerical values (ordinals or cardinals), dates, and/or units are changed. Modifiers or Adjectives: The modifiers and adjectives pertaining to a noun are changed. Negation or Antonyms: Introduce or remove a negation or change comparatives to their antonyms. Facts: Trivial factual errors not pertaining to the above classes are introduced in the translations. Dropped Content: A significant clause in the translation is removed. Please identify that error.\textbackslash{}nA: 
   \\
   \hline\noalign{\smallskip}
Movie Recommendation &
  F &
  Q: \{Options\}\textbackslash{}nFind a movie similar to \{Input\}\textbackslash{}nA: 
   \\
 &
  B* &
  Q: Find a movie similar to \{Input\}\textbackslash{}n\{Options\}\textbackslash{}nA: 
   \\
 &
  R &
  Q: \{Input\} Find a movie similar to\textbackslash{}n\{Options\}\textbackslash{}nA: \\
  \hline\noalign{\smallskip}
\end{tabular}
}
\caption{The position variants on BBH. * denote the reference position used in \citet{suzgun2022bbh}. In direct prompting under few-shot settings, additional instructions and examples are provided, along with extra rationales in CoT. Our prompts and inputs remain unchanged and adhere to \citet{srivastava2023beyond,suzgun2022bbh}, with modifications only on their relative position.}
\label{tab:reference_position_BBH}
\end{table*}


\section{All Results}
\label{appendix: all results}
\subsection{SST-2}
\label{appendix: SST-2}
The results of the 16-shot SST-2 preliminary experiment on T5-Large, as mentioned in Section \ref{gradient_free_setup}, are presented in Table \ref{tab:16-shot-sst2-preliminary}. We additionally calibrate the output distribution following the method outlined by \citet{zhao2021calibrate}, leading to enhanced in-context learning accuracy across all positions. Despite this improvement, the results continue to display a similar trend to our previous observations, including significant performance differences across various positions and a suboptimal reference prompt position.

 All other SST-2 results are presented in Tables \ref{tab:CM-SST-2}, \ref{tab:CC-SST-2}, \ref{tab:PM-SST-2}, \ref{tab:PC-SST-2}, \ref{tab:PC-XL-SST-2} and \ref{tab:zero-shot-flan-sst-2}.

\begin{table*}[ht!]
  \centering
  \resizebox{0.7\textwidth}{!}{
\begin{tabular}{lrrrr}
    \hline\noalign{\smallskip}
          & \multicolumn{1}{l}{\textbf{K=16}} & \multicolumn{1}{l}{\textbf{K=32}} & \multicolumn{1}{l}{\textbf{K=64}} & \multicolumn{1}{l}{\textbf{K=128}} \\
          & \multicolumn{1}{l}{\textbf{mean(std)}} & \multicolumn{1}{l}{\textbf{mean(std)}} & \multicolumn{1}{l}{\textbf{mean(std)}} & \multicolumn{1}{l}{\textbf{mean(std)}} \\
    \hline\noalign{\smallskip}
    \textbf{One prompt sequence} &       &       &       &  \\
    \hline\noalign{\smallskip}
    It was. \{text\_a\} {[}mask{]}  & \multicolumn{1}{l}{90.02(1.44)} & \multicolumn{1}{l}{91.15(1.95)} & \multicolumn{1}{l}{90.92(1.90)} & \multicolumn{1}{l}{91.81(0.99)} \\
    \{text\_a\} {[}mask{]} It was. & \multicolumn{1}{l}{88.62(1.69)} & \multicolumn{1}{l}{90.71(1.60)} & \multicolumn{1}{l}{91.31(1.69)} & \multicolumn{1}{l}{91.56(0.52)} \\
    \{text\_a\}. It was {[}mask{]}  & \multicolumn{1}{l}{90.21(1.58)} & \multicolumn{1}{l}{89.63(1.84)} & \multicolumn{1}{l}{91.74(1.37)} & \multicolumn{1}{l}{92.87(0.93)} \\
    \hdashline
    It was. {[}mask{]} \{text\_a\}  & \multicolumn{1}{l}{85.67(5.36)} & \multicolumn{1}{l}{89.15(1.79)} & \multicolumn{1}{l}{89.95(2.16)} & \multicolumn{1}{l}{91.28(1.04)} \\
    {[}mask{]} \{text\_a\} It was. & \multicolumn{1}{l}{85.92(3.55)} & \multicolumn{1}{l}{87.04(1.35)} & \multicolumn{1}{l}{90.30(1.34)} & \multicolumn{1}{l}{91.35(1.36)} \\
\end{tabular}%
    }
   \caption{Cloze manual prompt with RoBERTa-large on SST-2. The italic row indicates
the reference prompt position.}
  \label{tab:CM-SST-2}%
\end{table*}%

\begin{table*}[htb]
  \centering
    \resizebox{0.8\textwidth}{!}{
    \begin{tabular}{p{6cm}rrrr}
    \hline\noalign{\smallskip}
          & \multicolumn{1}{l}{\textbf{K=16}} & \multicolumn{1}{l}{\textbf{K=32}} & \multicolumn{1}{l}{\textbf{K=64}} & \multicolumn{1}{l}{\textbf{K=128}} \\
          & \multicolumn{1}{l}{\textbf{mean(std)}} & \multicolumn{1}{l}{\textbf{mean(std)}} & \multicolumn{1}{l}{\textbf{mean(std)}} & \multicolumn{1}{l}{\textbf{mean(std)}} \\
    \hline\noalign{\smallskip}
    \textbf{One prompt sequence} &       &       &       &  \\
    \hline\noalign{\smallskip}
    Question: Is this sentence positive or negative? Answer: \{text\_a\} & \multicolumn{1}{l}{85.73(6.94)} & \multicolumn{1}{l}{91.93(0.45)} & \multicolumn{1}{l}{\textbf{93.05(0.60)}} & \multicolumn{1}{l}{\textbf{93.39(0.39)}} \\
    \textit{\{text\_a\} Question: Is this sentence positive or negative?  Answer:} & \multicolumn{1}{l}{\textbf{89.45(4.16)}} & \multicolumn{1}{l}{\textbf{91.97(0.78)}} & \multicolumn{1}{l}{92.41(0.94)} & \multicolumn{1}{l}{93.17(0.45)} \\
    \hline\noalign{\smallskip}
    variance & 3.72  & 0.04  & 0.64  & 0.22 \\
    \hline\noalign{\smallskip}
    \textbf{Two prompt sequences} &       &       &       &  \\
    \hline\noalign{\smallskip}
    Question: Is this sentence positive or negative?  \{text\_a\} Answer:  & \multicolumn{1}{l}{85.94(10.00)} & \multicolumn{1}{l}{91.06(0.96)} & \multicolumn{1}{l}{92.41(0.94)} & \multicolumn{1}{l}{92.84(0.98)} \\
    \hline\noalign{\smallskip}
    variance all & 3.72  & 0.91  & 0.64  & 0.55 \\
    \hline\noalign{\smallskip}
    \end{tabular}%
    }
  \caption{Prefix manual prompt with T5-Large on SST-2. The italic row indicates the reference prompt position.}
\label{tab:PM-SST-2}%
\end{table*}%
\begin{table*}[htb]
  \centering
  \resizebox{0.7\textwidth}{!}{
    \begin{tabular}{lrrrr}
    \hline\noalign{\smallskip}
          & \multicolumn{1}{l}{\textbf{K=16}} & \multicolumn{1}{l}{\textbf{K=32}} & \multicolumn{1}{l}{\textbf{K=64}} & \multicolumn{1}{l}{\textbf{K=128}} \\
          & \multicolumn{1}{l}{\textbf{mean(std)}} & \multicolumn{1}{l}{\textbf{mean(std)}} & \multicolumn{1}{l}{\textbf{mean(std)}} & \multicolumn{1}{l}{\textbf{mean(std)}} \\
    \hline\noalign{\smallskip}
    \textbf{One prompt sequence} &       &       &       &  \\
    \hline\noalign{\smallskip}
    \textit{\textbf{\textit{P}} \{text\_a\} {[}mask{]}} & \multicolumn{1}{l}{86.12(3.28)} & \multicolumn{1}{l}{\textbf{89.72(0.59)}} & \multicolumn{1}{l}{90.18(0.80)} & \multicolumn{1}{l}{90.32(0.85)} \\
    \{text\_a\}  {[}mask{]}  \textbf{\textit{P}} & \multicolumn{1}{l}{82.64(4.62)} & \multicolumn{1}{l}{89.11(2.68)} & \multicolumn{1}{l}{90.50(0.68)} & \multicolumn{1}{l}{91.10(0.96)} \\
    \{text\_a\}  \textbf{\textit{P}}  {[}mask{]} & \multicolumn{1}{l}{72.09(6.69)} & \multicolumn{1}{l}{76.93(9.09)} & \multicolumn{1}{l}{83.23(3.27)} & \multicolumn{1}{l}{89.17(0.91)} \\
    \hdashline
    \textbf{\textit{P}}  {[}mask{]}  \{text\_a\} & \multicolumn{1}{l}{80.99(1.39)} & \multicolumn{1}{l}{82.00(4.65)} & \multicolumn{1}{l}{85.00(1.72)} & \multicolumn{1}{l}{86.90(2.75)} \\
    {[}mask{]}  \{text\_a\}  \textbf{\textit{P}} & \multicolumn{1}{l}{72.02(5.58)} & \multicolumn{1}{l}{71.83(5.71)} & \multicolumn{1}{l}{80.50(2.63)} & \multicolumn{1}{l}{84.98(1.31)} \\
\end{tabular}%
    }
  \caption{Cloze continuous prompt with RoBERTa-large on SST-2. The italic row indicates
the reference prompt position.}
  \label{tab:CC-SST-2}%
\end{table*}%

\begin{table*}[htb]
  \centering
   \resizebox{0.7\textwidth}{!}{
    \begin{tabular}{lrrrr}
    \hline\noalign{\smallskip}
          & \multicolumn{1}{l}{\textbf{K=16}} & \multicolumn{1}{l}{\textbf{K=32}} & \multicolumn{1}{l}{\textbf{K=64}} & \multicolumn{1}{l}{\textbf{K=128}} \\
          & \multicolumn{1}{l}{\textbf{mean(std)}} & \multicolumn{1}{l}{\textbf{mean(std)}} & \multicolumn{1}{l}{\textbf{mean(std)}} & \multicolumn{1}{l}{\textbf{mean(std)}} \\
    \hline\noalign{\smallskip}
    \textbf{One prompt sequence} &       &       &       &  \\
    \hline\noalign{\smallskip}
    \textit{\textbf{\textit{P}} \{text\_a\}}  & \multicolumn{1}{l}{57.20(6.23)} & \multicolumn{1}{l}{67.11(15.21)} & \multicolumn{1}{l}{83.76(4.90)} & \multicolumn{1}{l}{\textbf{88.53(3.67)}} \\
    \{text\_a\} \textbf{\textit{P}} & \multicolumn{1}{l}{\textbf{70.64(16.85)}} & \multicolumn{1}{l}{\textbf{75.48(14.73)}} & \multicolumn{1}{l}{\textbf{89.11(2.46)}} & \multicolumn{1}{l}{86.81(8.46)} \\
    \hline\noalign{\smallskip}
    variance & 13.44 &   8.37	&   5.35  &  	1.72
        \\
    \hline\noalign{\smallskip}
    \textbf{Two prompt sequences} &       &       &       &  \\
    \hline\noalign{\smallskip}
    \textbf{\textit{P}} \{text\_a\} \textbf{\textit{P}} & \multicolumn{1}{l}{56.90(3.14)} & \multicolumn{1}{l}{62.71(7.22)} & \multicolumn{1}{l}{75.05(9.16)} & \multicolumn{1}{l}{80.80(3.13)} \\
    \hline\noalign{\smallskip}
    variance all & 13.74 & 12.77 & 14.06 & 7.73 \\
    \hline\noalign{\smallskip}
    \end{tabular}%
    }
  \caption{Prefix continuous prompt with T5-Large on SST-2 Dataset. The italic row indicates the reference prompt position.}
  \label{tab:PC-SST-2}%
\end{table*}%

\begin{table*}[ht!]
\centering
\resizebox{0.7\textwidth}{!}{
\begin{tabular}{lllll}
 \hline\noalign{\smallskip}
& \textbf{K=16}      & \textbf{K=32}      & \textbf{K=64}      & \textbf{K=128}     \\
& \textbf{mean(std)} & \textbf{mean(std)} & \textbf{mean(std)} & \textbf{mean(std)} \\
 \hline\noalign{\smallskip}
\textbf{One prompt sequence}  &       &      &      &     \\
 \hline\noalign{\smallskip}
\textit{\textbf{P} \{text\_a\}}  & \textbf{84.93(6.43)}       & \textbf{90.57(2.32)}       & \textbf{92.82(1.01)}       & \textbf{93.33(1.39)}       \\
\{text\_a\} P                     & 78.85(15.79)      & 86.51(5.72)       & 91.51(0.80)        & 91.93(2.43)       \\
 \hline\noalign{\smallskip}
variance                      & 6.08  & 4.06 & 1.31 & 1.4 \\
 \hline\noalign{\smallskip}
\textbf{Two prompt sequences} &       &      &      &     \\
 \hline\noalign{\smallskip}
\textit{\textbf{P}} \{text\_a\} \textit{\textbf{P}} & 72.87(16.72)      & 88.33(4.32)       & 92.32(1.48)       & 92.11(0.97)       \\
 \hline\noalign{\smallskip}
variance all                  & 12.06 & 4.06 & 1.31 & 1.4 \\
 \hline\noalign{\smallskip}
\end{tabular}
}
\caption{Prefix continuous prompt with T5-XL on SST-2. The italic row indicates the reference prompt position.}
\label{tab:PC-XL-SST-2} 
\end{table*}

\begin{table*}[htbp!]
\centering
\resizebox{0.9\textwidth}{!}{
\begin{tabular}{llll}
  \hline\noalign{\smallskip}
 & PBF & ICL & C-ICL \\
 & mean(std) & mean(std) & mean(std)\\
   \hline\noalign{\smallskip}
Question: Is this sentence positive or negative?\textbackslash{}nAnswer:\textbackslash{}n\{text\_a\} & 82.94(12.69) & 61.79(14.77) & 72.11(13.85)\\
Question: Is this sentence positive or negative?\textbackslash{}n\{text\_a\}\textbackslash{}nAnswer: & 90.34(1.54)  & 55.99(14.56) & 60.64(12.31)\\
\textit{\{text\_a\}\textbackslash{}nQuestion: Is this sentence positive or negative?\textbackslash{}nAnswer: } & 89.36(3.64)  & 50.28(1.76) & 62.27(7.80) \\
  \hline\noalign{\smallskip}
\end{tabular}
}
\caption{16-shot preliminary experiments with T5-Large on SST-2. PBF denotes prompt-based fine-tuning, ICL denotes original in-context learning and C-ICL denotes calibrated in-context learning. The italic row indicates the reference prompt position.}
\label{tab:16-shot-sst2-preliminary}
\end{table*}
 
\begin{table*}[htbp!]
\centering
\small
\resizebox{0.4\textwidth}{!}{
\begin{tabular}{llll}
      \hline\noalign{\smallskip}
               & Front & Both  & Rear  \\
\hline\noalign{\smallskip}
Flan-T5-3B     & 95.30 & \textbf{95.41} & 95.30 \\
Flan-T5-11B    & 95.53 & 95.53 & 95.53 \\
Flan-LLaMA-13B & 94.95 & \textbf{95.07} & 94.38 \\
      \hline\noalign{\smallskip}
\end{tabular}
}
\caption{The accuracy of different prompt positions with Flan-based models on SST-2 under the zero-shot setting.}
\label{tab:zero-shot-flan-sst-2}
\end{table*}

\subsection{CR} \label{appendix: CR}
Full CR results can be found in Tables \ref{tab:CM-CR}, \ref{tab:CC-CR}, \ref{tab:PM-CR}, \ref{tab:PC-CR}, and \ref{tab:PC-XL-CR}.

\begin{table*}[htb!]
  \centering
   \resizebox{0.7\textwidth}{!}{
    \begin{tabular}{lrrrr}
    \hline\noalign{\smallskip}
          & \multicolumn{1}{l}{\textbf{K=16}} & \multicolumn{1}{l}{\textbf{K=32}} & \multicolumn{1}{l}{\textbf{K=64}} & \multicolumn{1}{l}{\textbf{K=128}} \\
          & \multicolumn{1}{l}{\textbf{mean(std)}} & \multicolumn{1}{l}{\textbf{mean(std)}} & \multicolumn{1}{l}{\textbf{mean(std)}} & \multicolumn{1}{l}{\textbf{mean(std)}} \\
    \hline\noalign{\smallskip}
    \textbf{One prompt sequence} &       &       &       &  \\
    \hline\noalign{\smallskip}
    It was. \{text\_a\} {[}mask{]}  & \multicolumn{1}{l}{89.86(1.63)} & \multicolumn{1}{l}{89.54(0.74)} & \multicolumn{1}{l}{\textbf{92.25(0.48)}} & \multicolumn{1}{l}{92.10(0.91)} \\
    \{text\_a\} {[}mask{]} It was. & \multicolumn{1}{l}{90.64(2.22)} & \multicolumn{1}{l}{90.76(0.35)} & \multicolumn{1}{l}{91.54(0.62)} & \multicolumn{1}{l}{92.19(0.83)} \\
    \{text\_a\}. It was {[}mask{]}  & \multicolumn{1}{l}{90.93(1.74)} & \multicolumn{1}{l}{\textbf{91.98(0.69)}} & \multicolumn{1}{l}{91.90(0.86)} & \multicolumn{1}{l}{92.06(0.88)} \\
    \hdashline
    It was. {[}mask{]} \{text\_a\}  & \multicolumn{1}{l}{90.24(1.65)} & \multicolumn{1}{l}{91.25(1.65)} & \multicolumn{1}{l}{91.91(0.84)} & \multicolumn{1}{l}{91.36(0.82)} \\
    {[}mask{]} \{text\_a\} It was. & \multicolumn{1}{l}{89.29(4.11)} & \multicolumn{1}{l}{89.96(2.39)} & \multicolumn{1}{l}{91.13(1.43)} & \multicolumn{1}{l}{91.78(0.42)} \\
\end{tabular}%
    }
  \caption{Cloze manual prompt with RoBERTa-large on CR. The italic row indicates the reference prompt position.}
  \label{tab:CM-CR}%
\end{table*}%

\begin{table*}[htb]
  \centering
   \resizebox{0.8\textwidth}{!}{
    \begin{tabular}{p{6cm}rrrr}
    \hline\noalign{\smallskip}
          & \multicolumn{1}{l}{\textbf{K=16}} & \multicolumn{1}{l}{\textbf{K=32}} & \multicolumn{1}{l}{\textbf{K=64}} & \multicolumn{1}{l}{\textbf{K=128}} \\
          & \multicolumn{1}{l}{\textbf{mean(std)}} & \multicolumn{1}{l}{\textbf{mean(std)}} & \multicolumn{1}{l}{\textbf{mean(std)}} & \multicolumn{1}{l}{\textbf{mean(std)}} \\
    \hline\noalign{\smallskip}
    \textbf{One prompt sequence} &       &       &       &  \\
    \hline\noalign{\smallskip}
    Question: Is this sentence positive or negative? Answer: \{text\_a\} & \multicolumn{1}{l}{89.98(0.38)} & \multicolumn{1}{l}{91.13(1.87)} & \multicolumn{1}{l}{93.06(1.03)} & \multicolumn{1}{l}{\textbf{93.91(0.38)}} \\
    \textit{\{text\_a\} Question: Is this sentence positive or negative?  Answer:} & \multicolumn{1}{l}{92.52(0.92)} & \multicolumn{1}{l}{92.87(0.45)} & \multicolumn{1}{l}{\textbf{93.70(0.34)}} & \multicolumn{1}{l}{93.70(0.63)} \\
    \hline\noalign{\smallskip}
    variance & 2.54  & 1.74  & 0.64  & 0.21 \\
    \hline\noalign{\smallskip}
    \textbf{Two prompt sequences} &       &       &       &  \\
    \hline\noalign{\smallskip}
    Question: Is this sentence positive or negative?  \{text\_a\} Answer:  & \multicolumn{1}{l}{\textbf{92.71(0.53)}} & \multicolumn{1}{l}{\textbf{92.97(0.58)}} & \multicolumn{1}{l}{93.52(0.67)} & \multicolumn{1}{l}{93.49(0.63)} \\
    \hline\noalign{\smallskip}
    variance all & 2.73  & 1.84  & 0.64  & 0.42 \\
    \hline\noalign{\smallskip}
    \end{tabular}%
    }
  \caption{Prefix manual prompt with T5-Large on CR. The italic row indicates the reference prompt position.}
  \label{tab:PM-CR}%
\end{table*}%

\begin{table*}[htb]
  \centering
  \resizebox{0.7\textwidth}{!}{
    \begin{tabular}{lrrrr}
    \hline\noalign{\smallskip}
          & \multicolumn{1}{l}{\textbf{K=16}} & \multicolumn{1}{l}{\textbf{K=32}} & \multicolumn{1}{l}{\textbf{K=64}} & \multicolumn{1}{l}{\textbf{K=128}} \\
          & \multicolumn{1}{l}{\textbf{mean(std)}} & \multicolumn{1}{l}{\textbf{mean(std)}} & \multicolumn{1}{l}{\textbf{mean(std)}} & \multicolumn{1}{l}{\textbf{mean(std)}} \\
    \hline\noalign{\smallskip}
    \textbf{One prompt sequence} &       &       &       &  \\
    \hline\noalign{\smallskip}
    \textit{\textbf{\textit{P}}  \{text\_a\}  {[}mask{]}} & \multicolumn{1}{l}{83.81(3.18)} & \multicolumn{1}{l}{88.97(0.99)} & \multicolumn{1}{l}{89.82(0.60)} & \multicolumn{1}{l}{90.74(0.94)} \\
    \{text\_a\}  {[}mask{]}  \textbf{\textit{P}} & \multicolumn{1}{l}{83.66(4.42)} & \multicolumn{1}{l}{88.92(0.93)} & \multicolumn{1}{l}{\textbf{91.10(0.93)}} & \multicolumn{1}{l}{91.72(0.79)} \\
    \{text\_a\}  \textbf{\textit{P}}  {[}mask{]} & \multicolumn{1}{l}{73.95(10.58)} & \multicolumn{1}{l}{83.60(3.08)} & \multicolumn{1}{l}{85.76(2.82)} & \multicolumn{1}{l}{88.37(1.10)} \\
    \hdashline
    \textbf{\textit{P}}  {[}mask{]}  \{text\_a\} & \multicolumn{1}{l}{80.18(2.87)} & \multicolumn{1}{l}{82.96(0.96)} & \multicolumn{1}{l}{84.24(2.03)} & \multicolumn{1}{l}{87.96(0.92)} \\
    {[}mask{]}  \{text\_a\}  \textbf{\textit{P}} & \multicolumn{1}{l}{71.81(4.54)} & \multicolumn{1}{l}{74.41(7.62)} & \multicolumn{1}{l}{86.29(2.33)} & \multicolumn{1}{l}{89.79(1.33)} \\
\end{tabular}%
    }
  \caption{Cloze continuous prompt with RoBERTa-large on CR. The italic row indicates the reference prompt position.}
  \label{tab:CC-CR}%
\end{table*}%

\begin{table*}[htbp!]
  \centering
   \resizebox{0.7\textwidth}{!}{
    \begin{tabular}{lrrrr}
    \hline\noalign{\smallskip}
          & \multicolumn{1}{l}{\textbf{K=16}} & \multicolumn{1}{l}{\textbf{K=32}} & \multicolumn{1}{l}{\textbf{K=64}} & \multicolumn{1}{l}{\textbf{K=128}} \\
          & \multicolumn{1}{l}{\textbf{mean(std)}} & \multicolumn{1}{l}{\textbf{mean(std)}} & \multicolumn{1}{l}{\textbf{mean(std)}} & \multicolumn{1}{l}{\textbf{mean(std)}} \\
    \hline\noalign{\smallskip}
    \textbf{One prompt sequence} &       &       &       &  \\
  \hline\noalign{\smallskip}
    \textit{\textbf{\textit{P}}  \{text\_a\}}  & \multicolumn{1}{l}{\textbf{86.77(1.97)}} & \multicolumn{1}{l}{\textbf{89.36(1.29)}} & \multicolumn{1}{l}{\textbf{92.02(0.81)}} & \multicolumn{1}{l}{\textbf{91.63(1.08)}} \\
    \{text\_a\}  \textbf{\textit{P}} & \multicolumn{1}{l}{77.57(4.27)} & \multicolumn{1}{l}{86.50(2.71)} & \multicolumn{1}{l}{90.60(0.96)} & \multicolumn{1}{l}{91.47(0.85)} \\
\hline\noalign{\smallskip}
    variance & 9.2   & 2.86  & 1.42  & 0.16 \\
 \hline\noalign{\smallskip}
    \textbf{Two prompt sequences} &       &       &       &  \\
\hline\noalign{\smallskip}
    \textbf{\textit{P}}  \{text\_a\}  \textbf{\textit{P}} & \multicolumn{1}{l}{60.37(6.49)} & \multicolumn{1}{l}{72.43(4.72)} & \multicolumn{1}{l}{78.68(5.92)} & \multicolumn{1}{l}{85.87(4.93)} \\
\hline\noalign{\smallskip}
    variance all & 26.4  & 16.93 & 13.34 & 5.76 \\
\hline\noalign{\smallskip}
    \end{tabular}%
    }
\caption{Prefix continuous prompt with T5-Large on CR. The italic row indicates the reference prompt position.}
\label{tab:PC-CR}%
\end{table*}%

\begin{table*}[ht!]
\centering
\resizebox{0.7\textwidth}{!}{
\begin{tabular}{lllll}
\hline\noalign{\smallskip}
& \textbf{K=16}      & \textbf{K=32}      & \textbf{K=64}      & \textbf{K=128}     \\
\textbf{}                         & \textbf{mean(std)} & \textbf{mean(std)} & \textbf{mean(std)} & \textbf{mean(std)} \\
\hline\noalign{\smallskip}
\textbf{One prompt sequence}      &                    &                    &                    &                    \\
\hline\noalign{\smallskip}
\textit{\textbf{P} \{text\_a\}}            & \textbf{86.70(1.47)}        & 89.97(2.95)       & 91.89(1.33)       & \textbf{93.86(0.55)}       \\
\{text\_a\} \textit{\textbf{P}}                    & 68.44(14.37)      & 88.31(6.24)       & 92.59(1.07)       & 93.81(0.55)       \\
\hline\noalign{\smallskip}
variance                          & 18.26              & 1.66               & 0.7                & 0.05               \\
\hline\noalign{\smallskip}
\textbf{Two prompt sequences}     &                    &                    &                    &                    \\
\hline\noalign{\smallskip}
\textit{\textbf{P}} \{text\_a\} \textit{\textbf{P}}  & 80.99(10.64)      & \textbf{90.37(1.50)}        & \textbf{92.73(0.97)}       & 93.37(0.81)       \\
\hline\noalign{\smallskip}
variance all                      & 18.26              & 2.06               & 0.84               & 0.49      \\ 
\hline\noalign{\smallskip}
\end{tabular}
}
\caption{Prefix continuous prompt with T5-XL on CR. The italic row indicates the reference prompt position.}
\label{tab:PC-XL-CR}
\end{table*}

\subsection{TREC} \label{appendix: TREC}
Full TREC results can be found in Tables \ref{tab:CM-TREC}, \ref{tab:CC-TREC}, \ref{tab:PM-TREC}, \ref{tab:PC-TREC} and \ref{tab:PC-XL-trec}.

\begin{table*}[htbp!]
\centering
\resizebox{0.9\textwidth}{!}{
\begin{tabular}{lllll}
\hline\noalign{\smallskip}
& \textbf{K=16}        & \textbf{K=32}      & \textbf{K=64}        & \textbf{K=128}       \\ 

& \textbf{mean(std)}   & \textbf{mean(std)} & \textbf{mean(std)}   & \textbf{mean(std)}   \\
\hline\noalign{\smallskip}
\textbf{One prompt sequence}                                          &                      &                    &                      &                      \\
\hline\noalign{\smallskip}
This question is related to category. \{text\_a\} {[}mask{]}          & 83.64(2.67)          & \textbf{88.80(1.80)} & 91.68(0.85)          & 93.98(0.52)          \\
\{text\_a\} {[}mask{]} This question is related to category.          & \textbf{85.68(2.46)} & 88.19(1.68)        & \textbf{92.29(0.33)} & \textbf{94.55(0.32)} \\
\{text\_a\} This question is related to category. {[}mask{]}          & 84.57(4.38)          & 88.66(1.85)        & 91.69(0.29)          & 94.07(0.41)          \\
    \hdashline
This question is related to category. {[}mask{]} \{text\_a\}          & 83.37(2.10)           & 87.22(1.18)        & 91.64(1.23)          & 93.86(0.83)          \\
{[}mask{]} \{text\_a\} This question is related to category.          & 82.50(4.32)           & 87.77(2.42)        & 91.57(0.75)          & 93.52(0.83)          \\
\end{tabular}
}
\caption{Cloze manual prompt with RoBERTa-large on TREC. The italic row indicates the reference prompt position.}
\label{tab:CM-TREC}
\end{table*}

\begin{table*}[htbp!]
\centering
\resizebox{0.85\textwidth}{!}{
\begin{tabular}{p{8cm}llll}
\hline\noalign{\smallskip}
& \textbf{K=16}        & \textbf{K=32}        & \textbf{K=64}        & \textbf{K=128}       \\ 
 & \textbf{mean(std)}   & \textbf{mean(std)}   & \textbf{mean(std)}   & \textbf{mean(std)}   \\
 \hline\noalign{\smallskip}
\textbf{One prompt sequence}                                                                                                         &                      &                      &                      &                      \\
\hline\noalign{\smallskip}
Answer: Categories: Abbreviation, Entity, Description, Person, Location, Quantity. What category best describes:\{text\_a\}          & \textbf{85.04(1.57)} & 88.14(0.76)          & 90.23(1.12)          & \textbf{92.16(0.82)} \\
\{text\_a\} Categories: Abbreviation, Entity, Description, Person, Location, Quantity. What category best describes: Answer:         & 84.22(1.27)          & \textbf{88.28(0.91)} & 90.07(1.08)          & 92.00(0.53)             \\
\hline\noalign{\smallskip}
variance     
& 0.82                 & 0.14                 & 0.16                 & 0.16                 \\
\hline\noalign{\smallskip}
\textbf{Two prompt sequences}    \\                               \hline\noalign{\smallskip}
\textit{Categories: Abbreviation, Entity, Description, Person, Location, Quantity. What category best describes:\{text\_a\} Answer:} & 84.65(1.78)          & 87.88(1.01)          & \textbf{90.52(1.05)} & 91.64(1.08)          \\
\hline\noalign{\smallskip}
variance all   
& 0.82                 & 0.4                  & 0.45                 & 0.52                 \\
\hline\noalign{\smallskip}
\end{tabular}
}
\caption{Prefix manual prompt with T5-Large on TREC. The italic row indicates the reference prompt position.}
\label{tab:PM-TREC}
\end{table*}

\begin{table*}[htbp!]
\centering
\resizebox{0.7\textwidth}{!}{
\begin{tabular}{lllll}
\hline\noalign{\smallskip}
& \textbf{K=16}        & \textbf{K=32}        & \textbf{K=64}        & \textbf{K=128}       \\ 
 & \textbf{mean(std)}   & \textbf{mean(std)}   & \textbf{mean(std)}   & \textbf{mean(std)}   \\
 \hline\noalign{\smallskip}
\textbf{One prompt sequence}                &                      &                      &                      &                      \\
\hline\noalign{\smallskip}
\textbf{\textit{P}}  \{text\_a\}  {[}mask{]}                 & \textbf{65.06(3.84)} & 71.89(1.63)          & 78.25(2.26)          & 86.76(2.18)          \\
\{text\_a\}  {[}mask{]}  \textbf{\textit{P}}                 & 56.29(3.20)          & 69.63(3.14)          & 78.39(4.26)          & 87.21(1.67)          \\
\{text\_a\}  \textbf{\textit{P}}  {[}mask{]}                 & 49.28(8.77)          & 66.09(4.44)          & 75.69(5.29)          & 84.76(0.87)          \\
\hdashline
\textit{\textbf{\textit{P}}  {[}mask{]}  \{text\_a\}}        & 58.26(3.28)          & 73.10(2.08)          & \textbf{81.76(2.65)} & 87.44(1.36)          \\
{[}mask{]}  \{text\_a\}  \textbf{\textit{P}}                 & 53.05(3.70)          & 64.84(2.61)          & 75.37(3.27)          & 83.96(4.43)          \\
\end{tabular}
}
\caption{Cloze continuous prompt with RoBERTa-large on TREC. The italic row indicates the reference prompt position.}
\label{tab:CC-TREC}
\end{table*}

\begin{table*}[htbp!]
\centering
\resizebox{0.7\textwidth}{!}{
\begin{tabular}{lllll}
\hline\noalign{\smallskip}
& \textbf{K=16}        & \textbf{K=32}        & \textbf{K=64}        & \textbf{K=128}       \\ 
\  & \textbf{mean(std)}   & \textbf{mean(std)}   & \textbf{mean(std)}   & \textbf{mean(std)}   \\
\hline\noalign{\smallskip}
\textbf{One prompt sequence}       &                      &                      &                      &                      \\
\hline\noalign{\smallskip}
\textit{\textbf{\textit{P}} \{text\_a\}}        & 69.43(1.71) & 74.79(2.28)          & 81.98(0.99)          & 87.75(1.06) \\
\{text\_a\} \textbf{\textit{P}}                 & 67.88(5.22)          & 77.84(2.24) & \textbf{85.51(1.09)} & 89.16(1.40)           \\
\hline\noalign{\smallskip}
variance                           & 1.55                 & 3.05                 & 3.53                 & 1.41                 \\
\hline\noalign{\smallskip}
\textbf{Two prompt sequences}      &                      &                      & \textbf{}            &                      \\
\hline\noalign{\smallskip}
\textbf{\textit{P}} \{text\_a\} \textbf{\textit{P}} & \textbf{71.78(2.34)} & \textbf{80.44(1.73)} & 85.46(1.53) & \textbf{89.41(1.24)} \\
\hline\noalign{\smallskip}
variance all                       & 3.9                  & 5.65                 & 3.53                 & 1.66                \\
\hline\noalign{\smallskip}
\end{tabular}
}
\caption{Prefix continuous prompt with T5-Large on TREC. The italic row indicates the reference prompt position.}
\label{tab:PC-TREC}
\end{table*}

\begin{table*}[ht!]
\centering
\resizebox{0.7\textwidth}{!}{
\begin{tabular}{lllll}
\hline\noalign{\smallskip}
                              & \textbf{K=16}        & \textbf{K=32}        & \textbf{K=64}        & \textbf{K=128}       \\
\textbf{} & \textbf{mean(std)} & \textbf{mean(std)} & \textbf{mean(std)} & \textbf{mean(std)} \\
\hline\noalign{\smallskip}
\textbf{One prompt sequence}  &                      &                      &                      &                      \\
\hline\noalign{\smallskip}
\textit{\textbf{P} \{text\_a\}}        & 75.33(1.85)          & 82.07(1.83)          & \textbf{88.58(1.19)} & 91.23(0.96)          \\
\{text\_a\} \textit{\textbf{P}}                & \textbf{80.77(2.79)} & \textbf{84.87(1.23)} & 88.21(1.28)          & 90.66(1.13)          \\
\hline\noalign{\smallskip}
variance                      & 5.44                 & 2.8                  & 0.37                 & 0.57                 \\
\hline\noalign{\smallskip}
\textbf{Two prompt sequences} &                      &                      &                      &                      \\
\hline\noalign{\smallskip}
\textit{\textbf{P}} \{text\_a\} \textit{\textbf{P}}               & 75.93(4.00)          & 82.59(2.20)          & 87.22(0.84)          & \textbf{91.74(0.47)} \\
\hline\noalign{\smallskip}
variance all                  & 5.44                 & 2.8                  & 1.36                 & 1.08    \\
\hline\noalign{\smallskip}
\end{tabular}
}
\caption{Prefix continuous prompt with T5-XL on TREC. The italic row indicates the reference prompt position.}
\label{tab:PC-XL-trec}
\end{table*}

\subsection{RTE}\label{appendix: RTE}
Full RTE results can be found in Tables \ref{tab:CM-RTE}, \ref{tab:CC-RTE}, \ref{tab:PM-RTE}, \ref{tab:PC-RTE} and \ref{tab:PC-XL-RTE}.

\begin{table*}[htbp!]
  \centering
  \resizebox{0.85\textwidth}{!}{
    \begin{tabular}{lrrrr}
    \hline\noalign{\smallskip}
          & \multicolumn{1}{l}{\textbf{K=16}} & \multicolumn{1}{l}{\textbf{K=32}} & \multicolumn{1}{l}{\textbf{K=64}} & \multicolumn{1}{l}{\textbf{K=128}} \\
          & \multicolumn{1}{l}{\textbf{mean(std)}} & \multicolumn{1}{l}{\textbf{mean(std)}} & \multicolumn{1}{l}{\textbf{mean(std)}} & \multicolumn{1}{l}{\textbf{mean(std)}} \\
    \hline\noalign{\smallskip}
    \textbf{One prompt sequence} &       &       &       &  \\
    \hline\noalign{\smallskip}
    Question: ? the Answer: . \{text\_a\} \{text\_b\} {[}mask{]} & \multicolumn{1}{l}{60.43(9.12)} & \multicolumn{1}{l}{65.99(3.36)} & \multicolumn{1}{l}{74.01(2.17)} & \multicolumn{1}{l}{78.19(1.85)} \\
    \{text\_a\} Question: ? the Answer: . \{text\_b\} {[}mask{]} & \multicolumn{1}{l}{60.87(7.46)} & \multicolumn{1}{l}{64.33(3.92)} & \multicolumn{1}{l}{73.00(2.39)} & \multicolumn{1}{l}{78.84(3.53)} \\
    \{text\_a\} \{text\_b\} Question: ? the Answer: . {[}mask{]} & \multicolumn{1}{l}{55.74(4.38)} & \multicolumn{1}{l}{59.86(5.55)} & \multicolumn{1}{l}{64.84(3.06)} & \multicolumn{1}{l}{76.39(1.95)} \\
    \{text\_a\} \{text\_b\} {[}mask{]} Question: ? the Answer: . & \multicolumn{1}{l}{57.04(5.36)} & \multicolumn{1}{l}{58.84(3.81)} & \multicolumn{1}{l}{68.16(1.50)} & \multicolumn{1}{l}{75.60(2.97)} \\
    \hdashline
    Question: ? the Answer: . \{text\_a\} {[}mask{]} \{text\_b\} & \multicolumn{1}{l}{68.88(5.46)} & \multicolumn{1}{l}{71.19(4.26)} & \multicolumn{1}{l}{74.66(2.51)} & \multicolumn{1}{l}{79.28(1.59)} \\
\end{tabular}%
    }
  \caption{Cloze manual prompt with RoBERTa-large on RTE. The italic row indicates the reference prompt position.}
  \label{tab:CM-RTE}%
\end{table*}%

\begin{table*}[htbp!]
  \centering
  \resizebox{0.75\textwidth}{!}{
    \begin{tabular}{lrrrr}
    \hline\noalign{\smallskip}
          & \multicolumn{1}{l}{\textbf{K=16}} & \multicolumn{1}{l}{\textbf{K=32}} & \multicolumn{1}{l}{\textbf{K=64}} & \multicolumn{1}{l}{\textbf{K=128}} \\
          & \multicolumn{1}{l}{\textbf{mean(std)}} & \multicolumn{1}{l}{\textbf{mean(std)}} & \multicolumn{1}{l}{\textbf{mean(std)}} & \multicolumn{1}{l}{\textbf{mean(std)}} \\
    \hline\noalign{\smallskip}
    \textbf{One prompt sequence} &       &       &       &  \\
    \hline\noalign{\smallskip}
    \textbf{\textit{P}}  \{text\_a\}  \{text\_b\}  {[}mask{]} & \multicolumn{1}{l}{52.13(4.78)} & \multicolumn{1}{l}{54.15(2.71)} & \multicolumn{1}{l}{50.04(3.32)} & \multicolumn{1}{l}{57.40(2.23)} \\
    \{text\_a\}  \textbf{\textit{P}}  \{text\_b\}  {[}mask{]} & \multicolumn{1}{l}{55.31(4.48)} & \multicolumn{1}{l}{57.69(1.54)} & \multicolumn{1}{l}{57.98(3.29)} & \multicolumn{1}{l}{58.12(2.27)} \\
    \{text\_a\}  \{text\_b\}  \textbf{\textit{P}}  {[}mask{]} & \multicolumn{1}{l}{51.12(2.38)} & \multicolumn{1}{l}{49.17(3.12)} & \multicolumn{1}{l}{51.05(2.40)} & \multicolumn{1}{l}{51.34(1.97)} \\
    \{text\_a\}  \{text\_b\}  {[}mask{]}  \textbf{\textit{P}}  & \multicolumn{1}{l}{49.39(2.69)} & \multicolumn{1}{l}{51.19(2.52)} & \multicolumn{1}{l}{51.62(2.99)} & \multicolumn{1}{l}{55.02(1.07)} \\
    \hdashline
    \textit{\textbf{\textit{P}}  \{text\_a\}  {[}mask{]}  \{text\_b\}} & \multicolumn{1}{l}{55.60(4.18)} & \multicolumn{1}{l}{59.13(1.00)} & \multicolumn{1}{l}{59.21(1.89)} & \multicolumn{1}{l}{60.14(1.01)} \\
\end{tabular}%
    }
  \caption{Cloze continuous prompt with RoBERTa-large on RTE. The italic row indicates the reference prompt position.}
  \label{tab:CC-RTE}%
\end{table*}%

\begin{table*}[htb]
  \centering
  \resizebox{0.9\textwidth}{!}{
    \begin{tabular}{lrrrr}
\hline\noalign{\smallskip}
          & \multicolumn{1}{l}{\textbf{K=16}} & \multicolumn{1}{l}{\textbf{K=32}} & \multicolumn{1}{l}{\textbf{K=64}} & \multicolumn{1}{l}{\textbf{K=128}} \\
          & \multicolumn{1}{l}{\textbf{mean(std)}} & \multicolumn{1}{l}{\textbf{mean(std)}} & \multicolumn{1}{l}{\textbf{mean(std)}} & \multicolumn{1}{l}{\textbf{mean(std)}} \\
\hline\noalign{\smallskip}
    \textbf{One prompt sequence} &       &       &       &  \\
\hline\noalign{\smallskip}
    Question: True or False? Answer: \{text\_a\} \{text\_b\} & \multicolumn{1}{l}{58.99(8.37)} & \multicolumn{1}{l}{\textbf{66.86(4.85)}} & \multicolumn{1}{l}{70.40(2.18)} & \multicolumn{1}{l}{74.73(1.77)} \\
    \{text\_a\} Question: True or False? Answer: \{text\_b\} & \multicolumn{1}{l}{\textbf{61.66(6.67)}} & \multicolumn{1}{l}{65.56(3.88)} & \multicolumn{1}{l}{70.69(1.60)} & \multicolumn{1}{l}{74.44(1.03)} \\
    \{text\_a\} \{text\_b\} Question: True or False? Answer:  & \multicolumn{1}{l}{55.45(2.78)} & \multicolumn{1}{l}{55.52(3.20)} & \multicolumn{1}{l}{60.00(4.47)} & \multicolumn{1}{l}{66.21(3.57)} \\
\hline\noalign{\smallskip}
    variance & 6.21  & 11.34 & 10.69 & 8.52 \\
\hline\noalign{\smallskip}
    \textbf{Two prompt sequences} &       &       &       &  \\
\end{tabular}%
    }
  \caption{Prefix manual prompt with T5-Large on RTE. The italic row indicates the reference prompt position.}
  \label{tab:PM-RTE}%
\end{table*}%

\begin{table*}[htb]
  \centering
  \resizebox{0.7\textwidth}{!}{
    \begin{tabular}{lrrrr}
    \hline\noalign{\smallskip}
          & \multicolumn{1}{l}{\textbf{K=16}} & \multicolumn{1}{l}{\textbf{K=32}} & \multicolumn{1}{l}{\textbf{K=64}} & \multicolumn{1}{l}{\textbf{K=128}} \\
          & \multicolumn{1}{l}{\textbf{mean(std)}} & \multicolumn{1}{l}{\textbf{mean(std)}} & \multicolumn{1}{l}{\textbf{mean(std)}} & \multicolumn{1}{l}{\textbf{mean(std)}} \\
    \hline\noalign{\smallskip}
    \textbf{One prompt sequence} &       &       &       &  \\
    \hline\noalign{\smallskip}
    \textit{\textbf{\textit{P}}  \{text\_a\}  \{text\_b\} } & \multicolumn{1}{l}{\textbf{52.06(4.80)}} & \multicolumn{1}{l}{52.35(3.22)} & \multicolumn{1}{l}{\textbf{57.33(2.88)}} & \multicolumn{1}{l}{59.13(1.18)} \\
    \{text\_a\}  \textbf{\textit{P}}  \{text\_b\}  & \multicolumn{1}{l}{48.16(1.95)} & \multicolumn{1}{l}{50.25(3.19)} & \multicolumn{1}{l}{50.47(3.02)} & \multicolumn{1}{l}{54.66(3.92)} \\
    \{text\_a\}  \{text\_b\}  \textbf{\textit{P}}  & \multicolumn{1}{l}{48.45(2.44)} & \multicolumn{1}{l}{52.13(4.53)} & \multicolumn{1}{l}{54.44(4.13)} & \multicolumn{1}{l}{\textbf{61.23(3.40)}} \\
    \hline\noalign{\smallskip}
    variance & 3.9   & 2.1   & 6.86  & 6.57 \\
    \hline\noalign{\smallskip}
    \textbf{Two prompt sequences} &       &       &       &  \\
\end{tabular}%
    }
  \caption{Prefix continuous prompt with T5-Large on RTE. The italic row indicates the reference prompt position.}
  \label{tab:PC-RTE}%
\end{table*}%

\begin{table*}[ht!]
\centering
\resizebox{0.7\textwidth}{!}{
\begin{tabular}{lllll}
\hline\noalign{\smallskip}
& \textbf{K=16} & \textbf{K=32} & \textbf{K=64} & \textbf{K=128} \\
\textbf{}                          & \textbf{mean(std)} & \textbf{mean(std)} & \textbf{mean(std)} & \textbf{mean(std)} \\
\hline\noalign{\smallskip}
\textbf{One prompt sequence}    &               &               &               &                \\
\hline\noalign{\smallskip}
\textit{\textbf{P} \{text\_a\} \{text\_b\}} & 51.4(1.5)          & \textbf{54.4(3.9)} & \textbf{62.3(2.9)} & \textbf{64.4(3.8)} \\
\{text\_a\} \textit{\textbf{P}} \{text\_b\}       & 52.3(4.2)     & 52.1(3.9)     & 55.9(4.9)     & 57.9(3.4)      \\
\{text\_a\} \{text\_b\} \textit{\textbf{P}}       & 48.4(1.8)     & 50.3(4.6)     & 51.9(3.2)     & 56.7(4.3)      \\
\hline\noalign{\smallskip}
variance                        & 3.9           & 4.1           & 10.4          & 7.7            \\
\hline\noalign{\smallskip}
\textbf{Two prompt sequences}   &               &               &               &                \\
\end{tabular}
}
\caption{Prefix continuous prompt with T5-XL on RTE. The italic row indicates the reference prompt position.}
\label{tab:PC-XL-RTE}
\end{table*}

\subsection{BoolQ} \label{appendix: Boolq}
Full BoolQ results can be found in Tables \ref{tab:CM-boolq}, \ref{tab:CC-boolq}, \ref{tab:PM-boolq}, \ref{tab:PC-boolq} and \ref{tab:PC-XL-Boolq}.

\begin{table*}[htb!]
  \centering
    \resizebox{0.85\textwidth}{!}{
    \begin{tabular}{lrrrr}
   \hline\noalign{\smallskip}
          & \multicolumn{1}{l}{\textbf{K=16}} & \multicolumn{1}{l}{\textbf{K=32}} & \multicolumn{1}{l}{\textbf{K=64}} & \multicolumn{1}{l}{\textbf{K=128}} \\
          & \multicolumn{1}{l}{\textbf{mean(std)}} & \multicolumn{1}{l}{\textbf{mean(std)}} & \multicolumn{1}{l}{\textbf{mean(std)}} & \multicolumn{1}{l}{\textbf{mean(std)}} \\
   \hline\noalign{\smallskip}
    \textbf{One prompt sequence} &       &       &       &  \\
  \hline\noalign{\smallskip}
    . Question: ? Answer: . \{text\_a\} \{text\_b\} {[}mask{]} & \multicolumn{1}{l}{64.14(4.13)} & \multicolumn{1}{l}{65.99(6.23)} & \multicolumn{1}{l}{71.93(2.04)} & \multicolumn{1}{l}{73.48(0.68)} \\
    \{text\_a\} . Question: ? Answer: . \{text\_b\} {[}mask{]} & \multicolumn{1}{l}{69.37(2.11)} & \multicolumn{1}{l}{68.48(3.12)} & \multicolumn{1}{l}{72.61(1.23)} & \multicolumn{1}{l}{74.07(1.13)} \\
    \{text\_a\} \{text\_b\} . Question: ? Answer: . {[}mask{]} & \multicolumn{1}{l}{67.30(3.53)} & \multicolumn{1}{l}{66.90(4.11)} & \multicolumn{1}{l}{71.31(3.61)} & \multicolumn{1}{l}{74.06(1.50)} \\
    \{text\_a\} \{text\_b\} {[}mask{]} . Question: ? Answer: . & \multicolumn{1}{l}{62.56(5.09)} & \multicolumn{1}{l}{64.80(4.51)} & \multicolumn{1}{l}{69.76(3.61)} & \multicolumn{1}{l}{72.98(2.08)} \\
    \hdashline
    . Question: ? Answer: . \{text\_a\} {[}mask{]} \{text\_b\} & \multicolumn{1}{l}{64.86(5.11)} & \multicolumn{1}{l}{68.75(3.15)} & \multicolumn{1}{l}{71.43(1.90)} & \multicolumn{1}{l}{74.97(0.36)} \\
\end{tabular}%
    }
  \caption{Cloze manual prompt with RoBERTa-large on BoolQ. The italic row indicates the reference prompt position.}
  \label{tab:CM-boolq}%
\end{table*}%

\begin{table*}[htb]
  \centering
  \resizebox{0.75\textwidth}{!}{
    \begin{tabular}{lrrrr}
  \hline\noalign{\smallskip}
          & \multicolumn{1}{l}{\textbf{K=16}} & \multicolumn{1}{l}{\textbf{K=32}} & \multicolumn{1}{l}{\textbf{K=64}} & \multicolumn{1}{l}{\textbf{K=128}} \\
          & \multicolumn{1}{l}{\textbf{mean(std)}} & \multicolumn{1}{l}{\textbf{mean(std)}} & \multicolumn{1}{l}{\textbf{mean(std)}} & \multicolumn{1}{l}{\textbf{mean(std)}} \\
\hline\noalign{\smallskip}
    \textbf{One prompt sequence} &       &       &       &  \\
\hline\noalign{\smallskip}
    \textbf{\textit{P}} \{text\_a\} \{text\_b\} {[}mask{]} & \multicolumn{1}{l}{53.38(5.90)} & \multicolumn{1}{l}{54.39(4.83)} & \multicolumn{1}{l}{60.29(1.97)} & \multicolumn{1}{l}{60.95(1.78)} \\
    \{text\_a\} \textbf{\textit{P}} \{text\_b\} {[}mask{]} & \multicolumn{1}{l}{53.58(8.90)} & \multicolumn{1}{l}{58.28(0.57)} & \multicolumn{1}{l}{56.45(0.99)} & \multicolumn{1}{l}{55.74(2.67)} \\
    \{text\_a\} \{text\_b\} \textbf{\textit{P}} {[}mask{]} & \multicolumn{1}{l}{48.99(6.59)} & \multicolumn{1}{l}{52.92(8.67)} & \multicolumn{1}{l}{53.67(6.24)} & \multicolumn{1}{l}{56.06(3.23)} \\
    \{text\_a\} \{text\_b\} {[}mask{]} \textbf{\textit{P}}  & \multicolumn{1}{l}{54.82(4.73)} & \multicolumn{1}{l}{53.09(7.90)} & \multicolumn{1}{l}{55.14(3.27)} & \multicolumn{1}{l}{54.53(4.30)} \\
    \hdashline
    \textit{\textbf{\textit{P}} \{text\_a\} {[}mask{]} \{text\_b\}} & \multicolumn{1}{l}{60.12(2.53)} & \multicolumn{1}{l}{57.43(5.76)} & \multicolumn{1}{l}{62.09(1.29)} & \multicolumn{1}{l}{60.39(2.47)} \\
\end{tabular}%
    }
  \caption{Cloze continuous prompt with RoBERTa-large on BoolQ. The italic row indicates the reference prompt position.}
  \label{tab:CC-boolq}%
\end{table*}%

\begin{table*}[htb!]
  \centering
    \resizebox{0.7\textwidth}{!}{
    \begin{tabular}{lrrrr}
    \hline\noalign{\smallskip}
          & \multicolumn{1}{l}{\textbf{K=16}} & \multicolumn{1}{l}{\textbf{K=32}} & \multicolumn{1}{l}{\textbf{K=64}} & \multicolumn{1}{l}{\textbf{K=128}} \\
          & \multicolumn{1}{l}{\textbf{mean(std)}} & \multicolumn{1}{l}{\textbf{mean(std)}} & \multicolumn{1}{l}{\textbf{mean(std)}} & \multicolumn{1}{l}{\textbf{mean(std)}} \\
      \hline\noalign{\smallskip}
    \textbf{One prompt sequence} &       &       &       &  \\
      \hline\noalign{\smallskip}
    Question: ? Answer: \{text\_a\} \{text\_b\}  & \multicolumn{1}{l}{57.54(7.27)} & \multicolumn{1}{l}{67.91(1.42)} & \multicolumn{1}{l}{72.61(0.69)} & \multicolumn{1}{l}{75.59(1.02)} \\
    \{text\_a\} Question: ? Answer: \{text\_b\}  & \multicolumn{1}{l}{60.91(2.83)} & \multicolumn{1}{l}{65.65(2.54)} & \multicolumn{1}{l}{72.29(0.66)} & \multicolumn{1}{l}{\textbf{76.54(1.30)}} \\
    \{text\_a\} \{text\_b\} Question: ? Answer:  & \multicolumn{1}{l}{61.33(4.71)} & \multicolumn{1}{l}{64.77(3.08)} & \multicolumn{1}{l}{70.73(1.78)} & \multicolumn{1}{l}{73.80(1.52)} \\
    \hline\noalign{\smallskip}
    variance & 3.79  & 3.14  & 1.88  & 2.74 \\
    \hline\noalign{\smallskip}
    \textbf{Two prompt sequences} &       &       &       &  \\
\end{tabular}%
    }
  \caption{Prefix manual prompt with T5-Large on BoolQ. The italic row indicates the reference prompt position.}
  \label{tab:PM-boolq}%
\end{table*}%

\begin{table*}[htb!]
  \centering
  \resizebox{0.7\textwidth}{!}{
    \begin{tabular}{lllll}
   \hline\noalign{\smallskip}
      & \textbf{K=16} & \textbf{K=32} & \textbf{K=64} & \textbf{K=128} \\
      \textbf{} & \textbf{mean(std)} & \textbf{mean(std)} & \textbf{mean(std)} & \textbf{mean(std)} \\
   \hline\noalign{\smallskip}
      \textbf{One prompt sequence} & & & & \\
    \hline\noalign{\smallskip}
      \textit{\textbf{\textit{P}}  \{text\_a\} \{text\_b\}} & 45.26(7.24) & 52.09(4.68) & 52.26(5.95) & 50.69(5.71) \\			
      \{text\_a\} \textbf{\textit{P}} \{text\_b\} & 47.41(6.26) & 50.43(8.51) & 52.73(4.86) & 53.88(5.39) \\
      \{text\_a\} \{text\_b\} \textbf{\textit{P}} & 47.09(6.65) & 51.63(3.99) & \textbf{52.85(4.58)} & 52.23(7.49) \\
      
       \hline\noalign{\smallskip}
      variance & 2.15 & 1.66 & 0.59 & 3.19 \\			
       \hline\noalign{\smallskip}
      \textbf{Two prompt sequences} & & & & \\
       \hline\noalign{\smallskip}
      \textbf{\textit{P}} \{text\_a\} \textbf{\textit{P}} \{text\_b\} & \textbf{49.55(8.25)} & \textbf{57.26(3.92)} & 50.14(4.24) & 51.42(6.06) \\	
   
      \textbf{\textit{P}} \{text\_a\} \{text\_b\} \textbf{\textit{P}} & 46.67(5.37) & 56.28(5.63) & 52.53(6.13) & 52.06(3.37) \\

      \{text\_a\} \textbf{\textit{P}} \{text\_b\} \textbf{\textit{P}} & 48.93(7.25) & 51.72(3.79) & 51.61(6.14) & 53.36(3.78) \\

       \hline\noalign{\smallskip}
      variance & 2.88 & 5.54 & 2.39 & 1.94 \\
      		
       \hline\noalign{\smallskip}
      \textbf{Three prompt sequences} & & & & \\
       \hline\noalign{\smallskip}
      \textbf{\textit{P}} \{text\_a\}  \textbf{\textit{P}} \{text\_b\} \textbf{\textit{P}} & 47.04(7.81) & 49.43(5.4) & 51.9(3.39) & \textbf{54.87(1.96)} \\

       \hline\noalign{\smallskip}
      variance all & 4.29 & 7.83 & 2.71 & 4.18 \\	

         \hline\noalign{\smallskip}
    \end{tabular}
  }
  \caption{Prefix continuous prompt with T5-Large on BoolQ. The italic row indicates the reference prompt position.}
  \label{tab:PC-boolq}
\end{table*}

\begin{table*}[ht!]
\centering
\resizebox{0.7\textwidth}{!}{
\begin{tabular}{lllll}
\hline\noalign{\smallskip}
& \textbf{K=16}         & \textbf{K=32}         & \textbf{K=64}         & \textbf{K=128}        \\
\textbf{}                                   & \textbf{mean(std)}    & \textbf{mean(std)}    & \textbf{mean(std)}    & \textbf{mean(std)}    \\
\hline\noalign{\smallskip}
\textbf{One prompt sequence}                &                       &                       &                       &                       \\
\hline\noalign{\smallskip}
\textit{\textbf{P} \{text\_a\} \{text\_b\}} & 47.02(9.07)          & 48.91(1.97)          & 52.63(3.98)          & 52.75(3.99)          \\
\{text\_a\} \textit{\textbf{P}} \{text\_b\}                   & \textbf{51.83(6.26)} & \textbf{53.72(4.75)} & 51.93(4.74)          & 54.59(6.81)          \\
\{text\_a\} \{text\_b\} \textit{\textbf{P}}                   & 45.85(6.68)          & 52.77(1.56)          & 54.17(4.98)          & 52.76(2.19)          \\
\hline\noalign{\smallskip}
variance                           & 5.98                  & 4.81                  & 2.24                  & 1.84                  \\
\hline\noalign{\smallskip}
\textbf{Two prompt sequences}               &                       &                       &                       &                       \\
\end{tabular}
}
\caption{Prefix continuous prompt with T5-XL on BoolQ. The italic row indicates the reference prompt position.}
\label{tab:PC-XL-Boolq}
\end{table*}

\subsection{BBH}\label{appendix: BBH}
For zero-shot, few-shot direct prompting, and few-shot chain-of-thought prompting, full results are shown in Table \ref{tab:full-zero-bbh}, Table \ref{tab:full-direct-bbh}, and Table \ref{tab:full-cot-bbh}.

\begin{table*}[ht!]
\centering
  \resizebox{\textwidth}{!}{
    \begin{tabular}{llll|lll|lll|lll|lll|lll}
    \hline\noalign{\smallskip}
      & \multicolumn{3}{c}{T5-3B} & \multicolumn{3}{c}{Flan-T5-3B} & \multicolumn{3}{c}{T5-11B} & \multicolumn{3}{c}{Flan-T5-11B} & \multicolumn{3}{c}{LLaMA-13B} & \multicolumn{3}{c}{Flan-LLaMA-13B} \\
                       & Front & Both & Rear & Front   & Both & Rear & Front  & Both & Rear & Front    & Both & Rear & Front & Both & Rear & Front      & Both & Rear \\
                       \hline\noalign{\smallskip}
Causal Judgement      & 44.4  & 48.7 & 46.5 & 59.9    & 63.1 & 62.6 & 50.3   & 48.7 & 48.7 & 57.2    & 59.9 & 60.4 & 50.8  & 47.6 & 48.1 & 58.3     & 59.4 & 57.8 \\
Disambiguation QA     & 31.6  & 33.2 & 32.4 & 68.4	& 66.4	& 68
   & 29.6   & 34.8 & 32.8 & 68.4	& 67.2	& 65.6   & 35.6 & 61.6 & 55.2
 & 57.2    & 63.2 & 61.2 \\
Sports Understanding  & 48    & 46   & 46   & 58.8	& 54.8	& 54.8
 & 48.8   & 51.2 & 51.6 & 69.2	& 68.8	& 69.2
 & 64.4  & 68.4 & 62.8 & 67.2     & 66.4 & 64.4 \\
Navigate               & 42    & 42   & 28   & 59.2	& 60.4	& 54.8
 & 47.6   & 44.8 & 40.4 & 60.4	& 58.8	& 61.6 & 58    & 58   & 58   & 57.2	& 58 & 56.8
 \\
Logical deduction (5)      & 22.4 & 22.4 & 22.4 & 42.4	& 46.8	& 50.4 & 23.2 & 20   & 18.4 & 50  &	55.2	 &55.2
 & 18 & 26 & 25.6 & 27.6 & 37.2 & 37.6
 \\
Logical deduction (7)      & 15.2 & 13.6 & 14   & 46.4	& 52.4	& 51.2
 & 17.6 & 18.4 & 18.8 & 55.6	& 58.8	& 59.2
 & 16 & 18.8	& 17.6	& 24.4	& 29.2	& 35.6
  \\
\end{tabular}
}
\caption{Full results via few-shot direct prompting between different positions on BBH}
\label{tab:full-direct-bbh}
\end{table*}

\begin{table*}[htbp!]
\centering
  \resizebox{\textwidth}{!}{
    \begin{tabular}{llll|lll|lll|lll|lll|lll}
    \hline\noalign{\smallskip}
      & \multicolumn{3}{c}{T5-3B} & \multicolumn{3}{c}{Flan-T5-3B} & \multicolumn{3}{c}{T5-11B} & \multicolumn{3}{c}{Flan-T5-11B} & \multicolumn{3}{c}{LLaMA-13B} & \multicolumn{3}{c}{Flan-LLaMA-13B} \\
                         & Front & Both & Rear & Front   & Both & Rear & Front  & Both & Rear & Front    & Both & Rear & Front & Both & Rear & Front      & Both & Rear \\
                         \hline\noalign{\smallskip}
Causal Judgement         & 42.8  & 43.3 & 42.8 & 54.5    & 57.2 & 60.4 & 47.6   & 48.1 & 45.5 & 57.2
  & 56.7 & 57.2 & 51.9  & 54   & 51.3 & 57.8       & 55.1 & 52.4 \\
Disambiguation QA        & 27.6  & 28.4 & 26   & 67.6    & 69.2 & 68.4 & 28     & 29.2 & 30   & 64.8     & 64   & 63.6 & 50    & 41.6 & 43.2 & 39.2       & 37.6 & 39.2 \\
Sports Understanding     & 46    & 46.8 & 45.6 & 56      & 52.8 & 58   & 55.2   & 55.2 & 59.2 & 61.2     & 59.6 & 64   & 76.8  & 78.4 & 78   & 77.2       & 74.4 & 76   \\
Navigate &
  27.6 &
  26.8 &
  22.4 &
  56.8 &
  56.4 &
  50 &
  28.4 &
  32.8 &
  30.8 &
  58.4 &
  61.6 &
  57.6 &
  59.6 &
  58.8 &
  56.8 &
  55.2 &
  57.6 &
  56.8 \\
Logical Deduction (5) &
  20.8 &
  16.8 &
  17.6 &
  27.6 &
  37.6 &
  32.8 &
  13.2 &
  20.4 &
  21.6 &
  42 &
  51.2 &
  49.2 &
  28.4 &
  29.6 &
  29.6 &
  29.6 &
  28.4 &
  32.8 \\
Logical Deduction (7) &
  14.4 &
  12.8 &
  11.2 &
  18.4 &
  30 &
  24 &
  10.8 &
  17.2 &
  15.2 &
  34.8 &
  53.6 &
  50 &
  22.8 &
  21.6 &
  19.6 &
  20 &
  19.6 &
  19.2 \\
\end{tabular}
}
\caption{Full results via few-shot CoT prompting between different positions on BBH}
\label{tab:full-cot-bbh}
\end{table*}

\begin{table*}[ht!]
\centering
\resizebox{0.8\textwidth}{!}{
    \begin{tabular}{l|lll|lll|lll|lll}
    \hline\noalign{\smallskip}
      & \multicolumn{3}{l}{Flan-T5-3B} & \multicolumn{3}{l}{Flan-T5-11B} & \multicolumn{3}{l}{LLaMA-13B} & \multicolumn{3}{l}{Flan-LLaMA-13B} \\
                         & Front   & Both & Rear & Front    & Both & Rear & Front & Both & Rear & Front      & Both & Rear \\
                           \hline\noalign{\smallskip}
Causal Judgement         & 59.9    & 61   & 61   & 59.9     & 61   & 61.5 & 53.5  & 51.9 & 50.8 & 58.8       & 55.6 & 53.5 \\
Disambiguation QA        & 69.2    & 66.8 & 68.8 & 64.8     & 68.8 & 64.8 & 32    & 52.8 & 37.6 & 43.6       & 60   & 58   \\
Sports Understanding     & 52      & 53.6 & 51.2 & 59.2     & 59.6 & 61.6 & 60.8  & 52   & 54.4 & 64.4       & 61.2 & 58.8 \\
Navigate                 & 57.6    & 58.4 & 60.8 & 60.4     & 60.8 & 59.2 & 42.4  & 42   & 49.6 & 58         & 58   & 59.2 \\
Logical Deduction (5)    & 46.4    & 50   & 50.4 & 49.6     & 55.2 & 56.8 & 18    & 21.2 & 19.6 & 22.8       & 38   & 36   \\
Logical Deduction (7)    & 47.2    & 52.4 & 53.2 & 54.4     & 61.2 & 62.4 & 14.8  & 14.8 & 14.8 & 18.4       & 33.6 & 32   \\
Logical Deduction (3)    & 60.4    & 64.8 & 64.8 & 69.6     & 75.2 & 71.2 & 32    & 36.4 & 39.6 & 42.8       & 49.6 & 50   \\
Penguins in a Table      & 40.4    & 41.1 & 32.2 & 47.9     & 41.8 & 38.4 & 15.8  & 26   & 25.3 & 23.3       & 31.5 & 22.6 \\
Salient Translation Err. & 39.2    & 42.4 & 40.4 & 50       & 50   & 50   & 18    & 14   & 13.6 & 19.6       & 24.8 & 22.8 \\
Movie Recommendation     & 38.4    & 54   & 67.2 & 40.4     & 46.4 & 55.2 & 21.6  & 56   & 46.4 & 29.6       & 65.6 & 67.6
\\
  \hline\noalign{\smallskip}
\end{tabular}
}
\caption{Full results via zero-shot prompting between different positions on BBH}
\label{tab:full-zero-bbh}
\end{table*}

\end{document}